\documentclass{article}

\usepackage[preprint]{neurips_2026}

\usepackage[utf8]{inputenc} 
\usepackage[T1]{fontenc}    
\usepackage[table]{xcolor}  
\usepackage[
    colorlinks=true,
    linkcolor=red,
    citecolor=blue!50,
    urlcolor=purple!50
]{hyperref}
\usepackage{url}            
\usepackage{booktabs}       
\usepackage{amsfonts}       
\usepackage{nicefrac}       
\usepackage{microtype}      
\usepackage{graphicx}       
\usepackage[most]{tcolorbox}
\usepackage{pifont}
\usepackage{multirow}

\usepackage{amsmath,amssymb,amsfonts}
\usepackage{tabularx}
\usepackage{geometry}
\usepackage{enumitem}
\usepackage{algorithm}
\usepackage{algpseudocode}
\usepackage{caption}

\definecolor{HeaderLight}{RGB}{235,242,250}
\definecolor{PanelLight}{RGB}{246,249,253}
\definecolor{MineLight}{RGB}{255,247,232}
\definecolor{HeaderText}{RGB}{36,76,125}
\definecolor{BestColor}{RGB}{178,74,42}
\definecolor{SecondColor}{RGB}{22,116,151}
\definecolor{EasyPurpleBg}{RGB}{242,235,250}
\definecolor{EasyPurpleText}{RGB}{105,55,155}
\newcommand{\best}[1]{\textbf{\textcolor{BestColor}{#1}}}
\newcommand{\second}[1]{\textcolor{SecondColor}{#1}}
\definecolor{lightbluebox}{RGB}{245,250,255}
\definecolor{blueframe}{RGB}{90,140,190}

\newtcolorbox{questionbox}{
  colback=lightbluebox,
  colframe=blueframe,
  boxrule=0.6pt,
  arc=4pt,
  left=7pt,
  right=7pt,
  top=2pt,
  bottom=2pt,
  width=\linewidth,
  fontupper=\itshape
}

\title{MineEvolve:\\
Self-Evolution with Accumulated Knowledge for Long-Horizon Embodied Minecraft Agents}

\author{Zhengwei Xie\thanks{Equal Contribution \hspace{0.5cm} $^\dagger$ Corresponding Author \\ \hspace{2pt} Mails: xiezhengwei0307@gmail.com, zhisheng.researcher@gmail.com}$^*$ \\ USTC \\
\And Zhisheng Chen$^*$ \\ NTU \\
\And Ziyan Weng$^*$ \\ CityU-DG \\
\And Jinhan Li \\ USTC \\
\And Chenglong Li \\ THU \\
\And Zikai Xiao \\ZJU \\
\And Jingwei Song \\ HKU \\
\And Jinhao Jing \\ CUHK-SZ \\
\And Vireo Zhang$^\dagger$ \\
\And Kun Wang$^\dagger$ \\ NTU \\
}

\begin{document}

\maketitle

\begin{figure*}[h]
  \centering
  \vspace{-0.8em}
  \includegraphics[width=\textwidth]{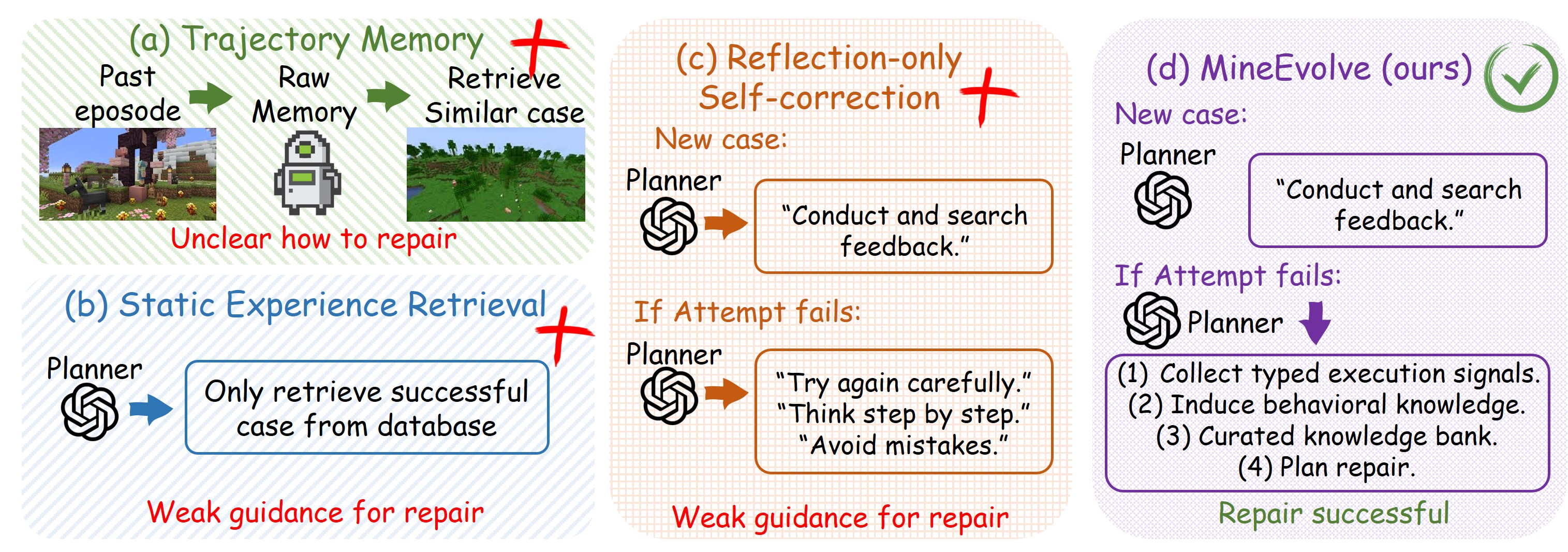}
   \vspace{-0.5em}
  \caption{
Prior methods either retrieve raw trajectories, reuse static successful cases, or rely on generic text reflections, which often provide weak guidance for repairing failures. In contrast, MineEvolve collects typed execution signals, induces actionable behavioral knowledge, maintains it in a curated knowledge bank, and uses it to repair the plan, turning past experience into explicit guidance for future decisions.
}
  \vspace{-0.8em}
\label{fig:diffrrence}
\end{figure*}

\begin{abstract}
\label{abstract}
Long-horizon embodied intelligence requires agents to improve through interaction, not merely to execute plans generated from static goals. A central challenge is therefore to transform past executions into knowledge that can shape future decisions. Minecraft provides a representative testbed for this problem, where tasks such as crafting tools, building redstone components, and obtaining diamond equipment involve long prerequisite chains and are frequently disrupted by missing tools, blocked paths, GUI failures, or stagnant execution. To this end, we propose \textbf{MineEvolve}, a knowledge-driven self-evolution framework that converts execution feedback into actionable behavioral knowledge. MineEvolve first uses \underline{\emph{\textbf{\ding{182}Monitor}}} to convert each subgoal execution into typed feedback, including state changes, inventory changes, failure types, progress signals, and stagnation indicators. \underline{\emph{\textbf{\ding{183}Inducer}}} then derives reusable skills from successful executions and remedies from failed or stagnant executions. \underline{\emph{\textbf{\ding{184}Curator}}} validates, merges, filters, and retrieves these knowledge entries, while \underline{\emph{\textbf{\ding{185}Adaptor}}} uses them to repair the unfinished part of the plan under repeated failures or stagnation. Experiments on the Minecraft MCU long-horizon task suite show that MineEvolve consistently improves performance across multiple language-model planners, with larger gains on high-dependency task groups. Ablation and knowledge-accumulation studies further demonstrate that converting execution signals into structured behavioral knowledge is an effective path toward self-evolving embodied agents in long-horizon environments. Our code is available at \url{https://github.com/xzw-ustc/MC-MineEvolve}.
\end{abstract}

\section{Introduction}
\label{introduction}

\begin{figure}[h]
    \centering
    \includegraphics[width=0.7\textwidth]{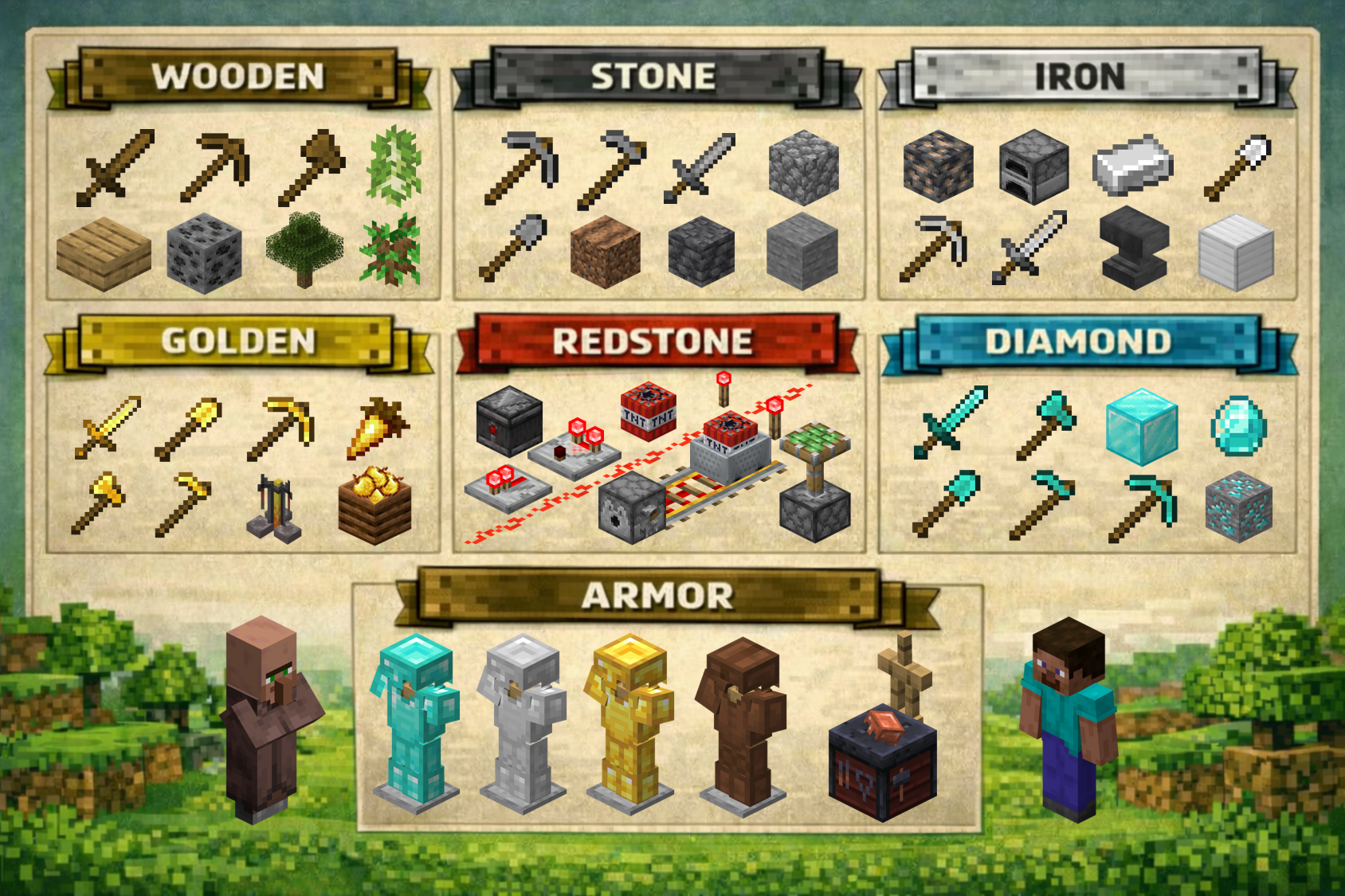}
    \caption{Progression of Minecraft from wooden to diamond tiers, 
illustrating the hierarchy of tools, materials, and equipment used to 
construct long-horizon tasks for embodied agent evaluation.}
    \label{fig:7task}
\end{figure}

Embodied agents are increasingly expected to perform long-horizon tasks in open and interactive environments, rather than merely follow short-horizon instructions~\citep{gupta2021embodied,liu2025embodied,luo2025large}. In such settings, an agent's ability depends not only on generating an initial plan from a goal, but also on converting past successes and failures into knowledge that can guide future behavior. Minecraft provides a representative test platform for this problem~\citep{johnson2016malmo,guss2019minerl,fan2022minedojo,zheng2025mcu}. Tasks such as crafting iron tools, building redstone components, 
and obtaining diamond equipment require long chains of prerequisites, including wood collection, basic tool crafting, ore search, smelting, crafting, and navigation. Failures in these tasks often arise not from a language model's misunderstanding of the goal, but from fine-grained execution errors, such as missing tools, blocked paths, abnormal GUI states, or omitted prerequisites. Although recent large language models have improved agents' ability to understand goals, decompose tasks, and use tools, a key question remains in long-dependency environments~\citep{brown2020language,achiam2023gpt,huang2022language,yao2022react,wang2026reasoning}:

\begin{questionbox}
How can an agent transform execution feedback into behavioral knowledge that is attributable, verifiable, and directly actionable for future planning?
\end{questionbox}

We address this question with \textbf{MineEvolve}, a knowledge-driven self-evolution framework for long-horizon embodied agents in Minecraft, which is achieved by continually constructing and using actionable knowledge from the agent's own executions. The framework follows four sequential steps: \underline{\emph{\textbf{\ding{182}Monitor}}} converts subgoal executions into typed execution feedback; \underline{\emph{\textbf{\ding{183}Inducer}}} derives skills from successful executions and remedies from failed or stagnant executions; \underline{\emph{\textbf{\ding{184}Curator}}} validates, merges, filters, and retrieves the generated knowledge; and \underline{\emph{\textbf{\ding{185}Adaptor}}} conditions planning on the retrieved knowledge and repairs the remaining plan when repeated failures or stagnation occur.

MineEvolve centers on converting \textit{execution feedback} into \textit{actionable knowledge}~\citep{shinn2023reflexion,zhao2024expel,madaan2023self}. During execution, state transitions, inventory changes, failure types, progress signals, and stagnation indicators provide structured information about the consequences of the agent's actions. MineEvolve organizes such feedback into two complementary forms of knowledge. Successful feedback is abstracted into skills, which encode preconditions, action sequences, and verification rules. Failed or stagnant feedback is transformed into remedies, which specify triggers, risk patterns, and repair actions. Skills support the reuse of effective behaviors, while remedies guide plan revision after local failures.

This design goes beyond trajectory memory and static experience retrieval. Rather than only retrieving similar past cases, MineEvolve identifies how past executions should modify the current plan~\citep{park2023generative,packer2023memgpt,lewis2020retrieval,kagaya2024rap}. For example, repeated movement near a target block without inventory gain can be converted into a remedy that suggests clearing blocking blocks or selecting an alternative route. A crafting failure caused by a missing crafting table can be converted into a prerequisite repair step before retrying the recipe. Thus, failures are not treated as passive records of unsuccessful attempts, but are converted into actionable knowledge that guides future plan revision.

As Figure~\ref{fig:7task}, we evaluate MineEvolve on the MCU long-horizon task suite~\citep{zheng2025mcu}. Across multiple language-model planners, MineEvolve consistently outperforms static experience retrieval baselines, with the largest improvements on high-dependency task groups such as Iron, Redstone, Diamond, and Armor. Our main contributions are summarized as follows:
\begin{itemize}
    \item \textbf{A knowledge-driven self-evolution framework for long-horizon Minecraft agents.}
    We introduce MineEvolve, which improves an agent through the continual accumulation and use of external behavioral knowledge. By grounding self-evolution in execution feedback, MineEvolve enables agents to convert past successes and failures into task knowledge that can guide future planning and local repair.

    \item \textbf{Feedback-conditioned generation of skills and remedies.}
    MineEvolve converts each execution into typed feedback through \textit{Monitor}, capturing state changes, inventory changes, failure types, progress signals, and stagnation indicators. Based on this feedback, \textit{Inducer} derives reusable skills from successful executions and remedies from failed or stagnant executions, producing knowledge with explicit triggers, verification rules, and repair actions.

    \item \textbf{Knowledge-guided repair of local failures.}
    MineEvolve maintains generated skills and remedies through \textit{Curator}, which validates, merges, filters, and retrieves knowledge relevant to the current context. The retrieved knowledge is then used by \textit{Adaptor} to revise only the remaining subgoals when repeated failures or stagnation occur, reducing ineffective retries and improving the stability of long-horizon task execution.
\end{itemize}

\section{Related Work}

\subsection{Minecraft Embodied Agents and Open-World Benchmarks}

Early research on Minecraft embodied agents mainly focused on reinforcement learning, imitation learning, and human demonstrations~\citep{johnson2016malmo,guss2019minerl,sutton1998reinforcement}. MineRL provides large-scale human demonstration data for studying sample-efficient learning in complex Minecraft tasks~\citep{guss2019minerl}. MineDojo further extends Minecraft into an open-ended embodied agent platform by integrating language-specified tasks, environment interaction, and internet-scale knowledge from videos, Wiki pages, and community content~\citep{fan2022minedojo}. For low-level behavior modeling, VPT learns visual-to-keyboard-and-mouse control policies from large-scale video pretraining~\citep{baker2022video}, while STEVE-1 instruction-tunes this paradigm to enable agents to follow short-horizon text or visual instructions~\citep{lifshitz2023steve}. These works provide important datasets, environments, and low-level control foundations for Minecraft agents, but they mainly focus on demonstration learning, behavior generation, or short-horizon instruction following, leaving long-dependency planning, failure recovery, and cross-task experience transformation less explored.

\subsection{LLM Planning and Embodied Control in Minecraft}

With the progress of large language models in task understanding and reasoning~\citep{brown2020language,achiam2023gpt,huang2022language,yao2022react}, Minecraft agents have increasingly adopted a hierarchical architecture with a high-level LLM planner and a low-level controller~\citep{ahn2022can,liang2023code,huang2022language}. DEPS improves subgoal generation and selection in open-world tasks through a Describe--Explain--Plan--Select pipeline~\citep{wang2023describe}. JARVIS-1 integrates visual observations, language instructions, task planning, and embodied control into a unified framework for multimodal Minecraft task execution~\citep{wang2024jarvis}. Recent work further strengthens the alignment between goals, observations, and actions. Optimus-2 combines multimodal LLM planning with a goal--observation--action conditioned policy, and uses an action-guided behavior encoder to improve embodied execution~\citep{li2025optimus-2}. These methods mainly address how to generate executable behaviors from language goals and visual observations, and substantially improve task decomposition, multimodal perception, and action execution~\citep{radford2021learning,liu2023visual}. However, long-horizon tasks also require agents to revise future plans based on execution feedback collected during interaction, which motivates research on experience memory and self-improvement.

\subsection{Experience Memory and Self-Improvement}

In Minecraft, Voyager uses an automatic curriculum, environment feedback, and a growing code skill library to accumulate transferable skills~\citep{wang2023voyager}. Optimus-1 introduces hybrid multimodal memory, organizing world knowledge as a hierarchical knowledge graph and storing historical experience summaries in a multimodal experience pool to support long-horizon planning and reflection~\citep{li2025optimus}. More recently, XENON treats long-horizon planning errors as external knowledge bias, and uses successful and failed experiences to revise dependency knowledge and action knowledge. These methods show that experience use in Minecraft agents is moving from trajectory retrieval and experience summarization~\citep{shinn2023reflexion,zhao2024expel,madaan2023self,park2023generative,packer2023memgpt} toward knowledge organization and knowledge revision~\citep{lewis2020retrieval,kagaya2024rap,parisi2019continual}. However, existing methods mostly focus on successful skill reuse, historical summaries, or dependency correction, while the conversion of fine-grained execution feedback into behavioral knowledge for plan repair remains underexplored. In contrast, MineEvolve focuses on transforming execution feedback into planning-time knowledge: each subgoal execution is first represented as typed feedback, then induced into skills or remedies, and finally used for knowledge-conditioned plan repair.

\section{Method}
\label{method}

\subsection{MineEvolve Overview}

\begin{figure*}[]
  \centering
  \vspace{-0.8em}
  \includegraphics[width=\textwidth]{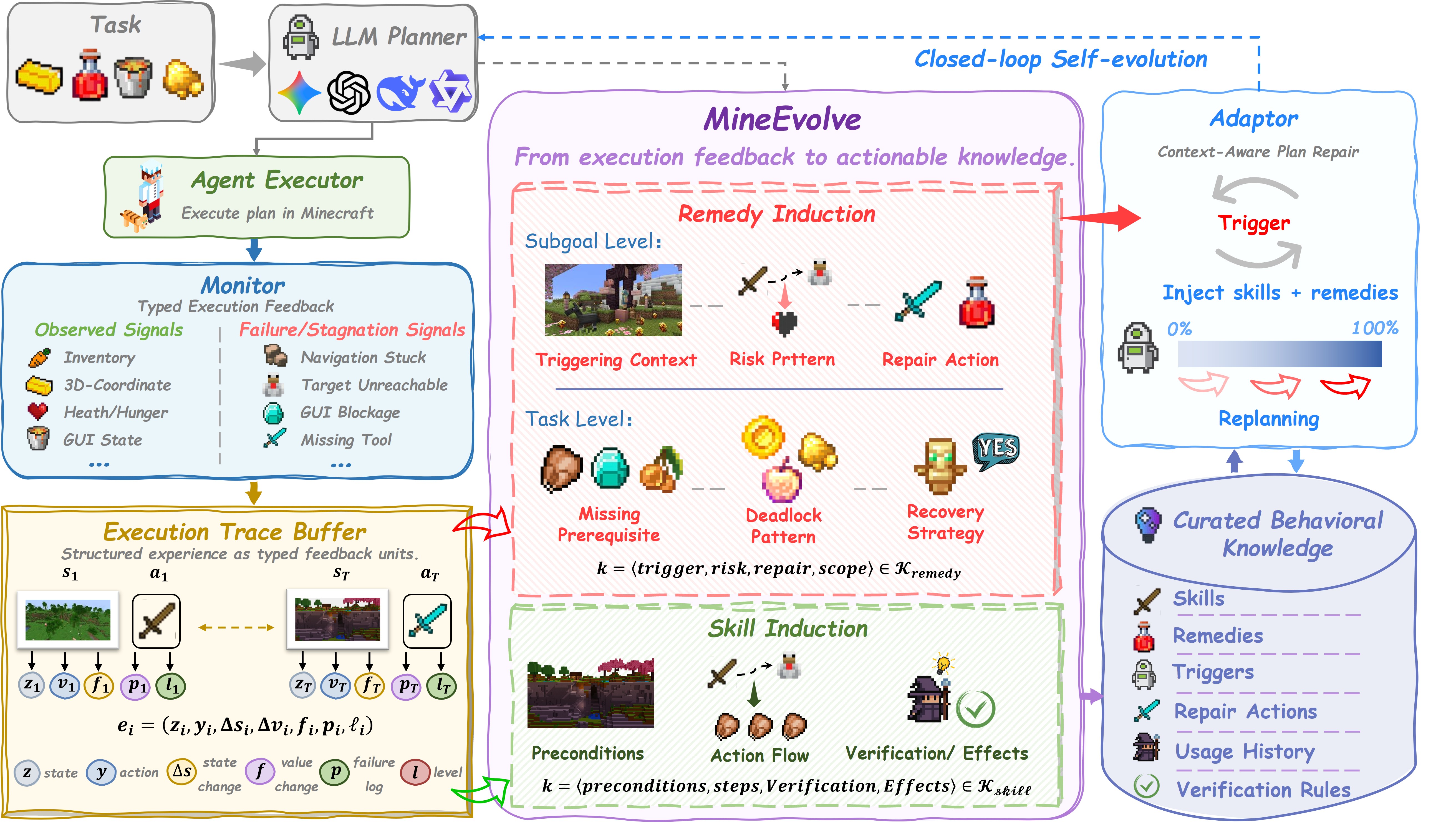}
   \vspace{-0.5em}
  \caption{
MineEvolve models self-evolution as the continual update and use of an external behavioral knowledge system, and consists of four sequential operations. First, \textit{\textbf{Monitor}} converts each subgoal execution into typed execution feedback. Second, \textit{\textbf{Inducer}} generates skills and remedies from this feedback. Third, \textit{\textbf{Curator}} validates, merges, filters, and retrieves candidate knowledge. Finally, \textit{\textbf{Adaptor}} injects the retrieved knowledge into the planner and repairs the unfinished part of the plan when repeated failures or stagnation occur.
}
  \vspace{-0.8em}
\label{fig:teaser}
\end{figure*}

As shown in Figure~\ref{fig:teaser}, MineEvolve targets long-horizon embodied tasks in Minecraft. Its goal is to help an agent accumulate and use external behavioral knowledge through interaction, without updating the parameters of the language model~\citep{packer2023memgpt,lewis2020retrieval,parisi2019continual}. Given a task goal $g$ and the current state $s_t$, the language-model planner generates a sequence of high-level subgoals, and a low-level executor carries them out in the environment. Instead of treating each attempt only as a success or failure, MineEvolve extracts typed execution feedback from each subgoal execution. It then turns successful feedback into \textit{skills} and failed or stagnant feedback into \textit{remedies}. Skills reuse verified successful procedures, while remedies describe risk patterns and repair actions.

\subsection{Feedback-Conditioned Skill and Remedy Generation}

For the $i$-th subgoal $z_i$, \textit{\textbf{Monitor}} compresses its execution into typed execution feedback:
\begin{equation}
e_i =
\big(
z_i,\;
y_i,\;
\Delta s_i,\;
\Delta v_i,\;
f_i,\;
p_i,\;
\ell_i
\big),
\end{equation}
where $y_i\in\{0,1\}$ denotes whether the subgoal succeeds, $\Delta s_i$ denotes the state change before and after execution, $\Delta v_i$ denotes the inventory change, $f_i$ denotes the failure type, $p_i$ denotes execution progress, and $\ell_i$ indicates stagnation or looping. Unlike free-form reflection~\citep{shinn2023reflexion,madaan2023self}, each field in this representation comes from environment states, inventory changes, or execution monitors, and can therefore be directly used by the knowledge generation process.

To capture failures of low-progress executions, Monitor combines movement, inventory change, and task-specific progress. We define the normalized progress of the $i$-th execution as
\begin{equation}
p_i =
\lambda_x
\frac{\operatorname{Var}(x_{i,1:T})}{\epsilon_x + \operatorname{Var}(x_{i,1:T})}
+
\lambda_v
\frac{\|\Delta v_i\|_1}{\epsilon_v+\|\Delta v_i\|_1}
+
\lambda_g p_i^{\mathrm{goal}},
\end{equation}
where $x_{i,1:T}$ is the coordinate sequence during the execution of $z_i$, $\Delta v_i$ is the inventory change, and $p_i^{\mathrm{goal}}$ is task-specific progress, such as crafting progress, GUI-state changes, or smelting progress. The coefficients $\lambda_x,\lambda_v,\lambda_g$ are weights. If progress is below a threshold and the subgoal is not completed, the execution is marked as stagnant:
\begin{equation}
\ell_i =
\mathbb{I}
\left[
p_i < \epsilon_p
\;\land\;
y_i = 0
\right].
\end{equation}
This rule allows the agent to detect low-progress behaviors such as repeated movement without effective approach, unchanged inventory, or GUI operations that produce no result.

Given typed execution feedback, \textit{\textbf{Inducer}} generates two forms of behavioral knowledge. A \textit{skill} is induced from successful feedback segments and describes the trigger context, preconditions, steps, verification rule, and observed effects for a task. A \textit{remedy} is induced from failed or stagnant feedback segments and describes the trigger context, failure type, risk pattern, repair action, and scope. Both are represented as a unified knowledge entry:
\begin{equation}
k =
\big(
\mathrm{type},\;
c,\;
u,\;
\phi,\;
\rho,\;
E
\big),
\qquad
\mathrm{type}\in\{\mathrm{skill},\mathrm{remedy}\}.
\end{equation}
Here, $c$ is the trigger context, $u$ is the knowledge content, $\phi$ is a verification or applicability condition, $\rho$ is the confidence score, and $E$ is the supporting set of execution feedback. For a skill, $u$ denotes reusable subgoal steps. For a remedy, $u$ denotes a risk pattern and the corresponding repair action. \textit{\textcolor{SecondColor}{The detailed pseudocode is shown in Appendix~\ref{algo2}.}}

When a feedback segment contains consecutive successful subgoals and its final state satisfies the corresponding verification rule, Inducer generates a skill. For example, if the agent successfully completes the sequence of collecting logs, crafting planks, and crafting sticks, and the target item appears in the inventory, the sequence can be induced as a reusable skill. In contrast, when failures repeat in a recent window or the current subgoal is stagnant, Inducer generates a remedy. The trigger is defined as
\begin{equation}
\frac{1}{h}
\sum_{j=i-h+1}^{i}
\mathbb{I}[y_j=0]
\geq
\eta_{\mathrm{fail}}
\quad
\text{or}
\quad
\ell_i=1,
\end{equation}
where $h$ is the recent execution window and $\eta_{\mathrm{fail}}$ is the failure-rate threshold. This condition prevents a single accidental failure from being immediately stored as a remedy, while still allowing repeated failure patterns or clear stagnation to be captured.

To prevent vague, conflicting, or non-executable text from entering the knowledge store, \textit{\textbf{Curator}} validates each candidate knowledge entry $k$. Validation includes schema completeness, context matchability, executability, specificity, and conflict filtering. We define the acceptance rule as
\begin{equation}
\mathrm{Accept}(k,\mathcal{K}) =
\mathbb{I}
\left[
V_{\mathrm{schema}}(k)
\land
V_{\mathrm{match}}(k)
\land
V_{\mathrm{exec}}(k)
\land
V_{\mathrm{spec}}(k)
\land
\neg C_{\mathrm{conflict}}(k,\mathcal{K})
\right],
\end{equation}
where $V_{\mathrm{schema}}$ checks whether all required fields are present, $V_{\mathrm{match}}$ checks whether the trigger can be matched to environment-state fields, $V_{\mathrm{exec}}$ checks whether the skill steps or repair action can be executed by the executor, $V_{\mathrm{spec}}$ rejects overly generic suggestions such as ``try again'' or ``be more careful'', and $C_{\mathrm{conflict}}$ checks whether the candidate conflicts with existing high-confidence knowledge. Only accepted candidates are written into the external knowledge store; otherwise, they remain only as feedback records for the current episode.

\subsection{Knowledge-guided repair of local failures.}

Curator maintains an external knowledge store $\mathcal{K}$. Each entry contains a trigger context, knowledge content, supporting feedback, confidence, support count, and usage history. Before planning, Curator retrieves relevant skills and remedies according to the current task goal, state, subgoal, inventory condition, spatial context, and recent failure types. To prevent the planning context from growing without bound, retrieval is constrained by a fixed prompt budget:
\begin{equation}
\mathcal{K}^{R}_i
=
\arg\max_{\mathcal{S}\subseteq \mathcal{K}}
\sum_{k\in\mathcal{S}} r(c_i,k),
\qquad
\mathrm{s.t.}
\quad
\sum_{k\in\mathcal{S}}\mathrm{tokens}(k)\leq B,
\end{equation}
where $c_i$ is the planning context at the $i$-th subgoal, $r(c_i,k)$ measures the relevance between knowledge entry $k$ and the current context, and $B$ is the prompt budget. Retrieved skills provide positive guidance for the planner, while remedies provide risk information and repair actions.

When repeated failures or stagnation occur, \textit{\textbf{Adaptor}} does not ask the language model to regenerate the entire plan. Instead, it keeps the completed prefix and repairs the unfinished part. Let the current plan be $\mathbf{z}_{1:N}$ and let the prefix $\mathbf{z}_{1:i}$ be completed or retained. Adaptor produces the repaired plan as
\begin{equation}
\mathbf{z}'_{1:N}
=
\left[
\mathbf{z}_{1:i},
\;
\mathrm{Repair}
\left(
g,\;
s_i,\;
\mathbf{z}_{1:i},\;
\mathcal{K}^{R}_i,\;
\mathcal{A}_i
\right)
\right],
\end{equation}
where $\mathcal{A}_i$ is the active set of remedies, including relevant remedies retrieved from the knowledge store and a temporary remedy induced from the current failure feedback. This design preserves useful progress while allowing new execution feedback to affect future planning. \textit{\textcolor{SecondColor}{The detailed pseudocode is shown in Appendix~\ref{algo3}.}}

When the agent repeatedly moves near a target block without any inventory change, MineEvolve identifies the pattern as a low-progress navigation failure and generates a remedy that asks the planner to avoid repeating the same path, clear blocking blocks, or choose an alternative route. When the agent fails to craft a target tool because a crafting table is missing, Adaptor inserts the prerequisite step of obtaining or placing a crafting table into the unfinished plan. Thus, MineEvolve does not simply ask the language model to ``try again''; it repairs the unfinished plan under explicit remedies derived from feedback, reducing ineffective retries and improving long-horizon execution stability.

\section{Experiments}
\label{sec:experiments}

\subsection{Experimental Setup}
\label{sec:exp_setup}
\paragraph{Environment and tasks.}
We evaluate MineEvolve on a 70-task subset selected from the Minecraft MCU tech-tree benchmark~\citep{zheng2025mcu}. The tasks span seven technology-tree groups, from early-stage Wooden, Stone, and Golden tasks to more difficult Iron, Redstone, Diamond, and Armor tasks with longer prerequisite chains. Agents operate in a survival-style environment with first-person RGB observations, structured state inputs, and an action space covering navigation, mining, placing, item usage, inventory control, and crafting. All episodes start with an empty inventory unless otherwise specified. We report Overall SR over all task groups and additionally focus on the hard groups, where prerequisite acquisition, recovery, and failure propagation are more prominent. \textit{\textcolor{SecondColor}{Detailed task-group statistics, representative tasks, and averaging rules are provided in Appendix~\ref{app:benchmark_details}.}}

\paragraph{Baselines and protocol control.}
We compare MineEvolve with DEPS, JARVIS-1, Optimus-1, and Optimus-2~\citep{wang2023describe,wang2024jarvis,li2025optimus,li2025optimus-2} under the same task set, task--seed split, episode horizon, and success criterion. DEPS, JARVIS-1, Optimus-1, and MineEvolve use STEVE-1 as the low-level execution policy~\citep{lifshitz2023steve}, while Optimus-2 uses its native GOAP-based policy. For STEVE-1-based methods, we additionally control the observation/action interface, action primitives, structured state fields, retrieval top-$K$, memory-token budget, and evaluation-time LLM-call budget.
\begin{table*}[t]
\centering
\caption{
Compact main results on the MCU task suite. 
Values denote success rate SR (\%). 
Easy Avg. is the task-count-weighted average over Wooden, Stone, and Gold tasks; 
Hard Avg. is the task-count-weighted average over Iron, Redstone, Diamond, and Armor tasks; 
Overall is the task-count-weighted average over all 70 tasks.
Best scores are highlighted in orange, and second-best scores are highlighted in teal.
}
\label{tab:main_results_panel}
\small
\setlength{\tabcolsep}{8pt}
\renewcommand{\arraystretch}{1.08}

\begin{tabular}{lccccc}
\toprule
\rowcolor{HeaderLight}
\textbf{Method}
& \textbf{Qwen3.5-Flash}
& \textbf{Qwen3.5-Plus}
& \textbf{GLM-4.7}
& \textbf{Gemini-3-Flash}
& \textbf{GPT-5.5} \\
\midrule

\rowcolor{EasyPurpleBg}
\multicolumn{6}{c}{\textbf{\textcolor{EasyPurpleText}{Weighted Easy Task Average}}} \\
DEPS             & 57.69 & 58.65 & 56.60 & 59.07 & 61.34 \\
JARVIS-1         & 70.47 & 71.04 & 69.30 & 70.62 & 72.41 \\
Optimus-1        & 73.76 & 73.66 & 72.55 & 73.26 & 75.29 \\
Optimus-2 (GOAP) & \second{73.93} & \second{74.50} & \best{73.28} & \second{74.86} & \second{75.60} \\
\rowcolor{MineLight}
\textbf{MineEvolve}
& \best{74.14}
& \best{75.61}
& \second{72.97}
& \best{75.47}
& \best{76.98} \\

\addlinespace[2pt]
\rowcolor{EasyPurpleBg}
\multicolumn{6}{c}{\textbf{\textcolor{EasyPurpleText}{Weighted Hard Task Average}}} \\
DEPS             &  9.71 &  9.66 &  8.82 & 10.06 & 11.92 \\
JARVIS-1         & 22.60 & 23.62 & 21.79 & 22.98 & 25.07 \\
Optimus-1        & 27.21 & 29.93 & 27.01 & 29.04 & 31.35 \\
Optimus-2 (GOAP) & \second{32.11} & \second{33.25} & \second{31.25} & \second{34.19} & \second{35.70} \\
\rowcolor{MineLight}
\textbf{MineEvolve}
& \best{34.06}
& \best{37.13}
& \best{32.06}
& \best{36.42}
& \best{40.23} \\

\addlinespace[2pt]
\rowcolor{EasyPurpleBg}
\multicolumn{6}{c}{\textbf{\textcolor{EasyPurpleText}{Weighted Overall}}} \\
DEPS             & 28.90 & 29.25 & 27.93 & 29.66 & 31.69 \\
JARVIS-1         & 41.75 & 42.59 & 40.79 & 42.04 & 44.01 \\
Optimus-1        & 45.83 & 47.42 & 45.23 & 46.73 & 48.92 \\
Optimus-2 (GOAP) & \second{48.84} & \second{49.75} & \second{48.06} & \second{50.45} & \second{51.66} \\
\rowcolor{MineLight}
\textbf{MineEvolve}
& \best{50.09}
& \best{52.52}
& \best{48.43}
& \best{52.04}
& \best{54.93} \\

\bottomrule
\end{tabular}
\end{table*}

\subsection{Main Experimental Results}
\label{sec:main_results}

Table~\ref{tab:main_results_panel} reports the main results on the MCU task suite. To highlight the benefits on long dependency chains, the main table reports three metrics: All aggregate metrics are computed at the task level rather than by assigning equal weight to task groups. 
Specifically, Easy Avg., Hard Avg., and Overall weight each group by the number of tasks it contains. \textit{\textcolor{SecondColor}{The complete results for all seven task groups are provided in Appendix Table~\ref{tab:app_full_results}.}}

Under the unified MC task protocol, MineEvolve achieves the highest Overall and Hard Avg. across multiple planner backbones. This trend suggests that the gains of MineEvolve do not depend on a single language model, but mainly arise from transforming execution feedback into behavioral knowledge and using such knowledge for conditioned local repair. The improvement is more pronounced on difficult task groups, indicating that remedies are particularly effective in repairing missing prerequisites, navigation stagnation, and crafting failures in long dependency chains.

\subsection{Controlled Ablation Study}
\label{sec:controlled_ablation}

To verify the source of the gains, we conduct targeted ablations using the Qwen3.5-Plus planner and the same STEVE-1 low-level execution policy. Each variant corresponds to one key design in MineEvolve: the Monitor provides typed execution feedback, the Inducer generates skills and remedies, the Curator verifies and maintains knowledge, and the Adaptor performs local repair. All variants use the same planner, low-level execution policy, interaction horizon, retrieval top-$K$, and evaluation-time LLM-call budget.

\begin{figure*}[!htbp]
\centering
\includegraphics[width=\linewidth]{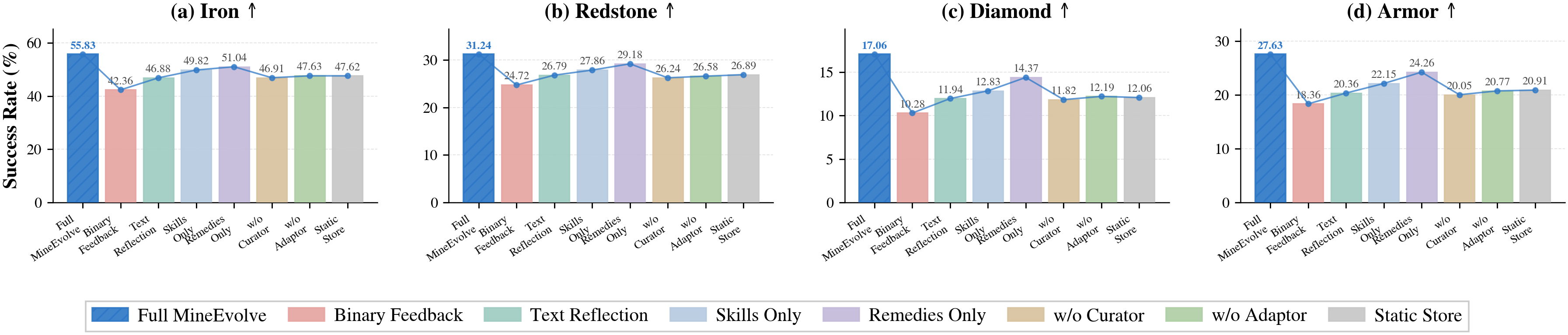}
\caption{Targeted ablation results on the four difficult task groups. Bars denote SR (\%). MineEvolve consistently outperforms ablated variants across Iron, Redstone, Diamond, and Armor tasks.}
\label{fig:targeted_ablation}
\end{figure*}

Figure~\ref{fig:targeted_ablation} shows the targeted ablation results on the four difficult task groups. Binary Feedback keeps only binary success/failure signals and is used to verify the necessity of typed execution feedback. Text Reflection replaces structured skills and remedies with free-form textual reflection. Skills Only and Remedies Only isolate the effects of reusing successful experience and failure-recovery knowledge, respectively. w/o Curator removes verification, merging, and conflict filtering. w/o Adaptor injects retrieved knowledge only during initial planning and does not perform local repair. Static Store freezes the knowledge base and is used to examine whether continuous knowledge updates truly bring self-evolutionary gains.

The ablation results show that each component contributes to performance on difficult tasks. Removing typed execution feedback causes the largest degradation across all four task groups, with Binary Feedback consistently performing worst. Replacing structured knowledge with free-form reflection improves over binary feedback but still remains substantially below the full system, suggesting that natural-language summaries alone are insufficient for reliable knowledge reuse. Skills Only and Remedies Only both improve over reflection, while the full system performs best, indicating that successful-experience reuse and failure-conditioned recovery knowledge are complementary. Remedies Only consistently outperforms Skills Only, suggesting that failure-conditioned recovery knowledge is especially important for difficult long-horizon tasks. Removing the Curator, the Adaptor, or continuous knowledge updates also leads to clear drops, showing that MineEvolve benefits not only from storing experience but also from verifying, updating, and applying it through local repair.

\begin{table}[!htbp]
\centering
\tiny
\caption{Ablation on feedback granularity. All variants share the same low-level execution policy, knowledge-base size, retrieval budget, replanning trigger conditions, and evaluation-time LLM-call budget. Values denote SR (\%).}
\label{tab:feedback_granularity}
\resizebox{\linewidth}{!}{
\begin{tabular}{lccccc}
\toprule
Source of Remedies & Iron & Redstone & Diamond & Armor & Hard Avg. \\
\midrule
Binary Feedback & 42.36 & 24.72 & 10.28 & 18.36 & 27.06 \\
Trajectory-level Reflection & 46.88 & 26.79 & 11.94 & 20.36 & 29.98 \\
Typed Execution Feedback & \textbf{55.83} & \textbf{31.24} & \textbf{17.06} & \textbf{27.63} & \textbf{37.13} \\
\bottomrule
\end{tabular}
}
\end{table}

Table~\ref{tab:feedback_granularity} further examines how remedy effectiveness depends on the granularity of execution feedback. Typed Execution Feedback substantially outperforms both binary feedback and trajectory-level reflection across all difficult task groups. This indicates that the effectiveness of remedies does not merely come from asking the LLM to summarize past trajectories, but from execution-feedback fields that are matchable, verifiable, and directly actionable during future planning and local repair.
\subsection{Knowledge Accumulation and Analysis}
\label{sec:analysis_main}

\begin{figure*}[!htbp]
\centering
\IfFileExists{fig/four_task_groups_side_by_side.png}{
    \includegraphics[width=\textwidth]{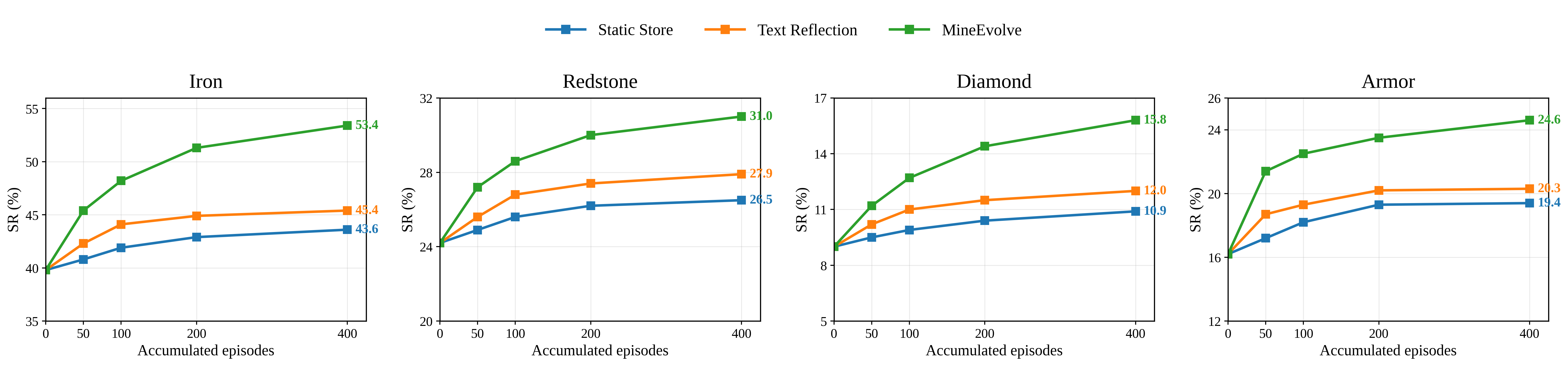}
}{
    \fbox{
    \begin{minipage}{0.95\textwidth}
    \vspace{1.1em}
    \centering
    \textbf{Placeholder: Group-wise Knowledge Accumulation Curves}\\
    \smallskip

    x-axis: accumulated episodes; y-axis: SR\\
    subplots: Iron, Redstone, Diamond, Armor\\
    curves: Static Store, Text Reflection, MineEvolve
    \vspace{1.1em}
    \end{minipage}}
}
\caption{Knowledge accumulation curves on the four difficult task groups. At each checkpoint, the external knowledge base is frozen before evaluation. Compared with Static Store and Text Reflection, MineEvolve shows more pronounced and sustained improvement on Iron, Redstone, Diamond, and Armor, indicating that structured skills and remedies can be accumulated and reused across task groups.}
\label{fig:experience_accumulation_groups}
\end{figure*}

To examine the self-evolving behavior of MineEvolve, we freeze the external knowledge base at different experience scales and evaluate on held-out hard tasks. The agent first interacts with the environment for $M$ episodes, and the knowledge base is fixed at checkpoints with $M\in\{0,50,100,200,400\}$. During evaluation, no further knowledge is updated or generated. This setting disentangles the gains from knowledge accumulation from online learning during evaluation. Figure~\ref{fig:experience_accumulation_groups} shows the knowledge accumulation curves on the four difficult task groups, i.e., Iron, Redstone, Diamond, and Armor, and \textit{\textcolor{SecondColor}{Appendix Table~\ref{tab:app_accumulation_by_group} reports the corresponding numerical results.}}

As shown in Figure~\ref{fig:experience_accumulation_groups}, all three methods improve to some extent on the four difficult task groups as the number of accumulated episodes increases, but the magnitude of improvement differs substantially. Static Store exhibits only limited growth, suggesting that relying solely on static retrieval or weakly updated knowledge is insufficient to fully exploit subsequent interaction experience. Text Reflection brings certain gains across task groups, but its improvement remains limited because free-form textual reflection lacks explicit triggering conditions, verification rules, and repair actions. In contrast, MineEvolve achieves more pronounced gains across all task groups, indicating that abstracting execution feedback into structured skills and remedies is better suited for sustained capability growth in long-horizon tasks.

The gains also differ reasonably across task groups. The Iron group starts and ends at relatively high performance, because its dependency chain is shorter than those of Diamond and Armor, and its tool-making and smelting procedures are easier to reuse. Diamond remains the most difficult task group throughout, but MineEvolve still improves steadily with accumulated experience, suggesting that remedies provide meaningful repair for sparse resources, tool prerequisites, and failures in underground search. The Armor group shows substantial improvement under MineEvolve, mainly benefiting from prerequisite completion, reuse of smelting procedures, and repair of crafting failures. The Redstone group has intermediate difficulty, and structured knowledge helps handle component crafting and multi-step dependencies.

In addition, \textit{\textcolor{SecondColor}{Appendix Table~\ref{tab:app_overhead}}} reports the knowledge-base size, retrieval tokens, retrieval latency, and the number of evaluation-time LLM calls, to verify that the budget control of the Curator can prevent unbounded context growth. \textit{\textcolor{SecondColor}{Appendix Section~\ref{app:diamond_curriculum}}} further analyzes the curriculum-style knowledge construction strategy on Diamond tasks, and \textit{\textcolor{SecondColor}{Appendix Section~\ref{app:case_study}}} provides qualitative case studies to illustrate how remedies transform failure feedback into plan repair.

\section{Conclusion}

We presented MineEvolve, a knowledge-driven self-evolution framework for long-horizon embodied agents in Minecraft. MineEvolve improves agents not by updating language-model parameters, but by continually converting execution feedback into actionable behavioral knowledge. Through monitor, inducer, curator, and adaptor, successful executions are distilled into reusable skills, while failed or stagnant executions are transformed into remedies for local plan repair. Experiments on the MCU long-horizon task suite show that MineEvolve consistently improves performance across multiple planner backbones, especially on high-dependency task groups such as Iron, Redstone, Diamond, and Armor. Ablations and knowledge-accumulation studies further confirm that typed feedback, skill/remedy generation, knowledge curation, and local repair are all important for sustained improvement. These results suggest that structured external behavioral knowledge provides an effective path toward self-evolving embodied agents in long-horizon environments.

\paragraph{Limitations.}
Despite its effectiveness, MineEvolve has several limitations.The current implementation is mainly evaluated in Minecraft and relies on environment-specific execution signals, such as inventory changes, GUI states, crafting progress, and failure types; transferring the framework to other embodied domains may require redesigning these feedback fields and verification rules.

\newpage

\bibliography{main}

@String(AAAI  = {AAAI})

@String(IJCAI = {IJCAI})

@article{brown2020language,
  title={Language Models are Few-Shot Learners},
  author={Brown, Tom and Mann, Benjamin and Ryder, Nick and Subbiah, Melanie and Kaplan, Jared D and Dhariwal, Prafulla and Neelakantan, Arvind and Shyam, Pranav and Sastry, Girish and Askell, Amanda and others},
  journal={Advances in neural information processing systems},
  volume={33},
  pages={1877--1901},
  year={2020}
}

@article{achiam2023gpt,
  title={GPT-4 Technical Report},
  author={Achiam, Josh and Adler, Steven and Agarwal, Sandhini and Ahmad, Lama and Akkaya, Ilge and Aleman, Florencia Leoni and Almeida, Diogo and Altenschmidt, Janko and Altman, Sam and Anadkat, Shyamal and others},
  journal={arXiv preprint arXiv:2303.08774},
  year={2023}
}

@article{gupta2021embodied,
  title={Embodied intelligence via learning and evolution},
  author={Gupta, Agrim and Savarese, Silvio and Ganguli, Surya and Fei-Fei, Li},
  journal={Nature communications},
  volume={12},
  number={1},
  pages={5721},
  year={2021},
  publisher={Nature Publishing Group UK London}
}

@inproceedings{johnson2016malmo,
  title={The Malmo Platform for Artificial Intelligence Experimentation},
  author={Johnson, Matthew and Hofmann, Katja and Hutton, Tim and Bignell, David},
  booktitle={Ijcai},
  volume={16},
  pages={4246--4247},
  year={2016}
}

@inproceedings{li2025optimus-2,
  title={Optimus-2: Multimodal minecraft agent with goal-observation-action conditioned policy},
  author={Li, Zaijing and Xie, Yuquan and Shao, Rui and Chen, Gongwei and Jiang, Dongmei and Nie, Liqiang},
  booktitle={Proceedings of the computer vision and pattern recognition conference},
  pages={9039--9049},
  year={2025}
}

@article{wang2023describe,
  title={Describe, explain, plan and select: Interactive planning with large language models enables open-world multi-task agents},
  author={Wang, Zihao and Cai, Shaofei and Chen, Guanzhou and Liu, Anji and Ma, Xiaojian and Liang, Yitao},
  journal={arXiv preprint arXiv:2302.01560},
  year={2023}
}

@article{lifshitz2023steve,
  title={Steve-1: A generative model for text-to-behavior in minecraft},
  author={Lifshitz, Shalev and Paster, Keiran and Chan, Harris and Ba, Jimmy and McIlraith, Sheila},
  journal={Advances in Neural Information Processing Systems},
  volume={36},
  pages={69900--69929},
  year={2023}
}

@inproceedings{yao2022react,
  title={ReAct: Synergizing Reasoning and Acting in Language Models},
  author={Yao, Shunyu and Zhao, Jeffrey and Yu, Dian and Du, Nan and Shafran, Izhak and Narasimhan, Karthik R and Cao, Yuan},
  booktitle={The eleventh international conference on learning representations},
  year={2022}
}

@inproceedings{zheng2025mcu,
  title={MCU: An evaluation framework for open-ended game agents},
  author={Zheng, Xinyue and Lin, Haowei and He, Kaichen and Wang, Zihao and Fu, Qiang and Fu, Haobo and Zheng, Zilong and Liang, Yitao},
  booktitle={Forty-second International Conference on Machine Learning},
  year={2025}
}

@article{guss2019minerl,
  title={Minerl: A large-scale dataset of minecraft demonstrations},
  author={Guss, William H and Houghton, Brandon and Topin, Nicholay and Wang, Phillip and Codel, Cayden and Veloso, Manuela and Salakhutdinov, Ruslan},
  journal={arXiv preprint arXiv:1907.13440},
  year={2019}
}

@article{shinn2023reflexion,
  title={Reflexion: language agents with verbal reinforcement learning},
  author={Shinn, Noah and Cassano, Federico and Gopinath, Ashwin and Narasimhan, Karthik and Yao, Shunyu},
  journal={Advances in neural information processing systems},
  volume={36},
  pages={8634--8652},
  year={2023}
}

@inproceedings{zhao2024expel,
  title={ExpeL: LLM Agents Are Experiential Learners},
  author={Zhao, Andrew and Huang, Daniel and Xu, Quentin and Lin, Matthieu and Liu, Yong-Jin and Huang, Gao},
  booktitle={Proceedings of the AAAI Conference on Artificial Intelligence},
  volume={38},
  number={17},
  pages={19632--19642},
  year={2024}
}

@article{wang2023voyager,
  title={Voyager: An Open-Ended Embodied Agent with Large Language Models},
  author={Wang, Guanzhi and Xie, Yuqi and Jiang, Yunfan and Mandlekar, Ajay and Xiao, Chaowei and Zhu, Yuke and Fan, Linxi and Anandkumar, Anima},
  journal={arXiv preprint arXiv:2305.16291},
  year={2023}
}

@article{wang2024jarvis,
  title={JARVIS-1: Open-World Multi-Task Agents With Memory-Augmented Multimodal Language Models},
  author={Wang, Zihao and Cai, Shaofei and Liu, Anji and Jin, Yonggang and Hou, Jinbing and Zhang, Bowei and Lin, Haowei and He, Zhaofeng and Zheng, Zilong and Yang, Yaodong and others},
  journal={IEEE Transactions on Pattern Analysis and Machine Intelligence},
  volume={47},
  number={3},
  pages={1894--1907},
  year={2024},
  publisher={IEEE}
}

@article{li2025optimus,
  title={Optimus-1: Hybrid Multimodal Memory Empowered Agents Excel in Long-Horizon Tasks},
  author={Li, Zaijing and Xie, Yuquan and Shao, Rui and Chen, Gongwei and Jiang, Dongmei and Nie, Liqiang},
  journal={Advances in neural information processing systems},
  volume={37},
  pages={49881--49913},
  year={2025}
}

@article{liu2025embodied,
  title={Embodied intelligence: A synergy of morphology, action, perception and learning},
  author={Liu, Huaping and Guo, Di and Cangelosi, Angelo},
  journal={ACM Computing Surveys},
  volume={57},
  number={7},
  pages={1--36},
  year={2025},
  publisher={ACM New York, NY}
}

@article{liu2023visual,
  title={Visual instruction tuning},
  author={Liu, Haotian and Li, Chunyuan and Wu, Qingyang and Lee, Yong Jae},
  journal={Advances in neural information processing systems},
  volume={36},
  pages={34892--34916},
  year={2023}
}

@article{fan2022minedojo,
  title={Minedojo: Building open-ended embodied agents with internet-scale knowledge},
  author={Fan, Linxi and Wang, Guanzhi and Jiang, Yunfan and Mandlekar, Ajay and Yang, Yuncong and Zhu, Haoyi and Tang, Andrew and Huang, De-An and Zhu, Yuke and Anandkumar, Anima},
  journal={Advances in Neural Information Processing Systems},
  volume={35},
  pages={18343--18362},
  year={2022}
}

@article{baker2022video,
  title={Video pretraining (vpt): Learning to act by watching unlabeled online videos},
  author={Baker, Bowen and Akkaya, Ilge and Zhokov, Peter and Huizinga, Joost and Tang, Jie and Ecoffet, Adrien and Houghton, Brandon and Sampedro, Raul and Clune, Jeff},
  journal={Advances in Neural Information Processing Systems},
  volume={35},
  pages={24639--24654},
  year={2022}
}

@inproceedings{radford2021learning,
  title={Learning transferable visual models from natural language supervision},
  author={Radford, Alec and Kim, Jong Wook and Hallacy, Chris and Ramesh, Aditya and Goh, Gabriel and Agarwal, Sandhini and Sastry, Girish and Askell, Amanda and Mishkin, Pamela and Clark, Jack and others},
  booktitle={International conference on machine learning},
  pages={8748--8763},
  year={2021},
  organization={PmLR}
}

@article{packer2023memgpt,
  title={MemGPT: towards LLMs as operating systems.},
  author={Packer, Charles and Fang, Vivian and Patil, Shishir\_G and Lin, Kevin and Wooders, Sarah and Gonzalez, Joseph\_E},
  year={2023},
  publisher={ArXiv}
}

@article{madaan2023self,
  title={Self-refine: Iterative refinement with self-feedback, 2023},
  author={Madaan, Aman and Tandon, Niket and Gupta, Prakhar and Hallinan, Skyler and Gao, Luyu and Wiegreffe, Sarah and Alon, Uri and Dziri, Nouha and Prabhumoye, Shrimai and Yang, Yiming and others},
  journal={URL https://arxiv. org/abs/2303.17651},
  volume={2303},
  year={2023}
}

@inproceedings{park2023generative,
  title={Generative agents: Interactive simulacra of human behavior},
  author={Park, Joon Sung and O'Brien, Joseph and Cai, Carrie Jun and Morris, Meredith Ringel and Liang, Percy and Bernstein, Michael S},
  booktitle={Proceedings of the 36th annual acm symposium on user interface software and technology},
  pages={1--22},
  year={2023}
}

@article{ahn2022can,
  title={Do as i can, not as i say: Grounding language in robotic affordances},
  author={Ahn, Michael and Brohan, Anthony and Brown, Noah and Chebotar, Yevgen and Cortes, Omar and David, Byron and Finn, Chelsea and Fu, Chuyuan and Gopalakrishnan, Keerthana and Hausman, Karol and others},
  journal={arXiv preprint arXiv:2204.01691},
  year={2022}
}

@article{luo2025large,
  title={Large language model agent: A survey on methodology, applications and challenges},
  author={Luo, Junyu and Zhang, Weizhi and Yuan, Ye and Zhao, Yusheng and Yang, Junwei and Gu, Yiyang and Wu, Bohan and Chen, Binqi and Qiao, Ziyue and Long, Qingqing and others},
  journal={arXiv preprint arXiv:2503.21460},
  year={2025}
}

@article{lewis2020retrieval,
  title={Retrieval-augmented generation for knowledge-intensive nlp tasks},
  author={Lewis, Patrick and Perez, Ethan and Piktus, Aleksandra and Petroni, Fabio and Karpukhin, Vladimir and Goyal, Naman and K{\"u}ttler, Heinrich and Lewis, Mike and Yih, Wen-tau and Rockt{\"a}schel, Tim and others},
  journal={Advances in neural information processing systems},
  volume={33},
  pages={9459--9474},
  year={2020}
}

@article{wang2026reasoning,
  title={Why Reasoning Fails to Plan: A Planning-Centric Analysis of Long-Horizon Decision Making in LLM Agents},
  author={Wang, Zehong and Wu, Fang and Wang, Hongru and Tang, Xiangru and Li, Bolian and Yin, Zhenfei and Ma, Yijun and Li, Yiyang and Sun, Weixiang and Chen, Xiusi and others},
  journal={arXiv preprint arXiv:2601.22311},
  year={2026}
}

@inproceedings{huang2022language,
  title={Language models as zero-shot planners: Extracting actionable knowledge for embodied agents},
  author={Huang, Wenlong and Abbeel, Pieter and Pathak, Deepak and Mordatch, Igor},
  booktitle={International conference on machine learning},
  pages={9118--9147},
  year={2022},
  organization={PMLR}
}

@article{kagaya2024rap,
  title={Rap: Retrieval-augmented planning with contextual memory for multimodal llm agents},
  author={Kagaya, Tomoyuki and Yuan, Thong Jing and Lou, Yuxuan and Karlekar, Jayashree and Pranata, Sugiri and Kinose, Akira and Oguri, Koki and Wick, Felix and You, Yang},
  journal={arXiv preprint arXiv:2402.03610},
  year={2024}
}

@article{parisi2019continual,
  title={Continual lifelong learning with neural networks: A review},
  author={Parisi, German I and Kemker, Ronald and Part, Jose L and Kanan, Christopher and Wermter, Stefan},
  journal={Neural networks},
  volume={113},
  pages={54--71},
  year={2019},
  publisher={Elsevier}
}

@inproceedings{liang2023code,
  title={Code as policies: Language model programs for embodied control},
  author={Liang, Jacky and Huang, Wenlong and Xia, Fei and Xu, Peng and Hausman, Karol and Ichter, Brian and Florence, Pete and Zeng, Andy},
  booktitle={2023 IEEE International conference on robotics and automation (ICRA)},
  pages={9493--9500},
  year={2023},
  organization={IEEE}
}

@book{sutton1998reinforcement,
  title={Reinforcement learning: An introduction},
  author={Sutton, Richard S and Barto, Andrew G and others},
  volume={1},
  number={1},
  year={1998},
  publisher={MIT press Cambridge}
}
\bibliographystyle{plainnat}


\appendix
\newpage
\section{Game Introduction}
Minecraft (MC) is an open-ended 3D sandbox game where players interact with a procedurally generated world composed of discrete blocks. From the perspective of agent research, MC naturally induces long-horizon sequential decision making: under \textbf{partial observability}, an agent must explore, gather resources, craft tools, and plan multi-step actions to accomplish goals. Its highly combinatorial state space and flexible objectives make Minecraft a widely used testbed for studying embodied control, planning, and tool-use behaviors in complex environments.

\subsection{Basic Rules}
\begin{figure}[htpb]
	\centering
	\includegraphics[width=0.95\linewidth]{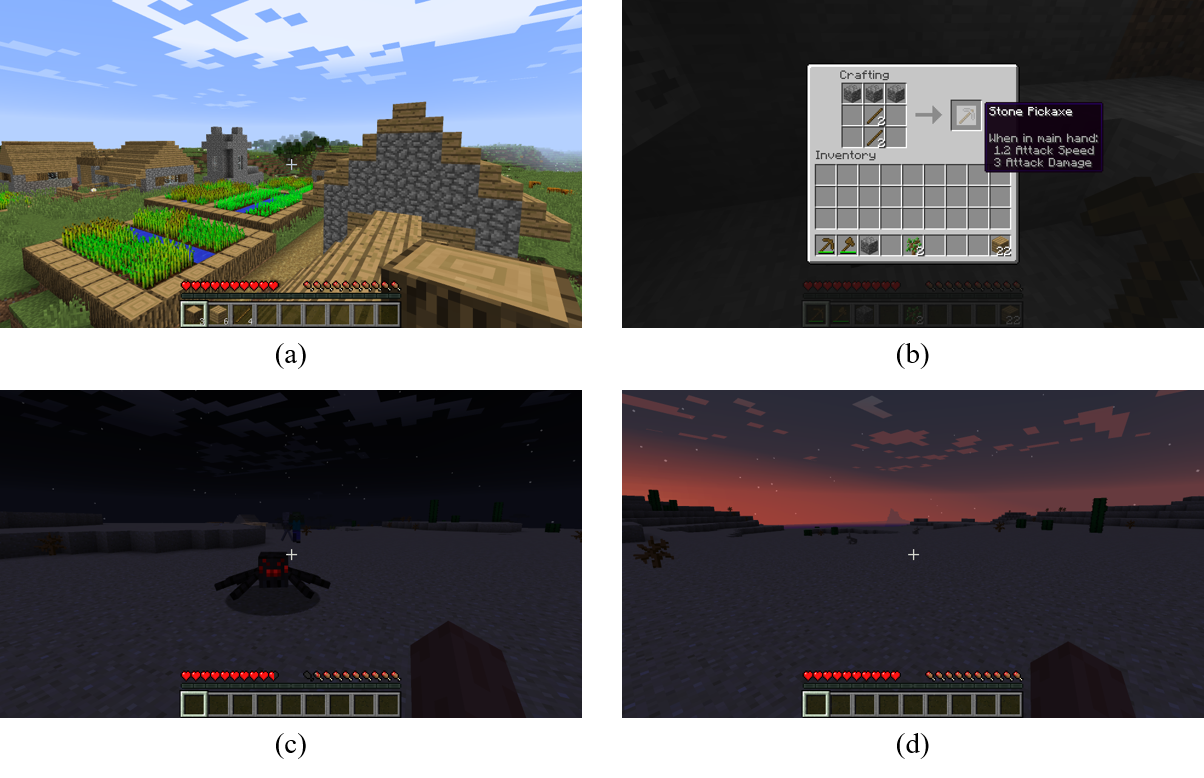}
	\caption{(a) Diverse block types form an explorable, harvestable, and modifiable world. (b) The crafting interface illustrates recipe-based item composition and tool progression. (c) Survival mode introduces continuous survival pressure from health, hunger, and environmental hazards. (d) The day--night cycle changes visibility and threats, making exploration and shelter building more strategic.}
	\label{fig:mc-basic-rules}
\end{figure}

A Minecraft world is composed of discrete blocks and populated with resources (e.g., wood, stone, ores), passive animals, and hostile mobs depending on the selected mode and difficulty. Many tasks follow a characteristic progression of \textbf{collecting} materials, \textbf{crafting} intermediate items, \textbf{upgrading} equipment, and \textbf{unlocking} access to more advanced resources or regions. Figure~\ref{fig:mc-basic-rules} provides representative in-game screenshots that illustrate these mechanics and the typical interaction context.

\begin{itemize}
	\item \textbf{Block interaction}: Agents can mine blocks to obtain items and place blocks to construct structures or modify the terrain; tool choice affects mining speed and obtainable drops.
	\item \textbf{Resource acquisition and crafting}: Collected items can be combined via crafting recipes to produce tools, weapons, and utilities; the crafting graph induces clear multi-step dependencies.
	\item \textbf{Survival dynamics (optional)}: In Survival mode, health and hunger constrain behavior; damage from falls, lava, or hostile mobs introduces risk and requires reactive control.
	\item \textbf{Time and environment}: Day--night cycles and biomes change visibility and threats, making navigation and shelter building strategic.
\end{itemize}

\newpage
\section{Compute Reporting}
\label{app:cpt}
\subsection{Experimental Configuration}
\label{app:exp_config}

All experiments are conducted on a server equipped with 8 NVIDIA A40 GPUs. The GPUs are primarily used for parallel Minecraft environment interaction, visual observation processing, and inference of the STEVE-1 low-level execution policy. The high-level planner, Monitor, Inducer, Curator, and Adaptor are executed under the unified protocol described in the main text, with the same evaluation-time LLM-call budget across all compared methods and ablation variants.

To ensure a fair comparison, all methods share the same task set, episode horizon, task--seed split, success criterion, and evaluation metric. For methods that use STEVE-1 as the low-level execution policy, we further fix the observation/action interface, action primitives, structured state fields, retrieval top-$K$, memory-token budget, and knowledge retrieval budget.

The main computational cost comes from environment rollouts, low-level policy inference, knowledge retrieval, and high-level LLM calls. Since this work focuses on the accumulation, validation, and reuse of external behavioral knowledge rather than parameter updates of the language model or the low-level policy, we do not train the planner backbone or STEVE-1. This setup ensures that the reported performance differences mainly reflect differences in planning, feedback utilization, and local repair mechanisms, rather than differences in hardware resources or training compute.
\subsection{Environment, Tasks, and Protocol Control}
\label{app:env_protocol}

We evaluate MineEvolve on a 70-task evaluation suite selected from the Minecraft MCU tech-tree benchmark. Agents operate in a survival-style environment and perceive the world through first-person RGB observations together with structured state inputs. The structured state includes the agent's position, health, hunger, GUI status, inventory contents, and observable task progress. The action space covers navigation, camera control, mining, attacking, placing blocks, using items, switching inventory slots, and crafting. Each episode starts with an empty inventory, and all task-required resources must be obtained through interaction with the environment.

The selected 70 tasks are grouped according to the Minecraft tech-tree progression into seven categories: Wooden, Stone, Iron, Gold, Redstone, Diamond, and Armor. The early-stage groups mainly involve basic resource collection and simple crafting, while the later-stage groups require longer dependency chains, including tool upgrades, ore search, smelting, redstone-component construction, diamond-equipment crafting, and armor crafting. To report the overall success rate across all task groups, we separately analyze the Iron, Redstone, Diamond, and Armor groups, as these tasks more clearly expose missing prerequisites, local failure propagation, and the need for recovery planning.

We define the \textbf{low-level execution policy} as the model that maps observations and high-level subgoals to executable Minecraft actions. Our main comparison follows a system-level evaluation protocol: all methods are evaluated on the same task set, task--seed split, episode horizon, success criterion, and success-rate metric, while each method retains its native high-level planning and execution pipeline. DEPS, JARVIS-1, Optimus-1, and MineEvolve use STEVE-1 as the low-level execution policy under our interface. Optimus-2, in contrast, uses its native GOAP-based execution policy and is included as a system-level baseline with a different low-level execution design.

For STEVE-1-based methods, we further control the action primitives, observation/action interface, structured state fields, memory-token budget, retrieval top-$K$, and evaluation-time LLM-call budget. This ensures that the same-policy comparisons are not confounded by differences in the low-level controller, input representation, retrieval budget, or LLM-call budget. Typed execution feedback is constructed from the same environment states and execution-monitoring signals across all relevant methods. However, only MineEvolve further abstracts this feedback into reusable skills and remedies, which are then used for knowledge-conditioned local repair during planning and execution.

\newpage
\section{Algorithmic Details}
\label{app:algorithm_details}

The main text describes the core design of MineEvolve: the agent performs self-evolution through an external behavioral knowledge system rather than by updating language-model parameters. Specifically, MineEvolve converts each subgoal execution into typed execution feedback, induces skills and remedies from the feedback, validates and maintains the generated knowledge, and performs knowledge-conditioned local replanning when repeated failures or stagnation occur. This appendix provides full pseudocode for the procedure.

\subsection{Overall MineEvolve Loop}
\label{algo1}

Algorithm~\ref{alg:app_mineevolve_loop} shows the overall loop of MineEvolve. The system maintains an episode-level feedback buffer $\mathcal{B}$ and an external knowledge store $\mathcal{K}$ containing validated skills and remedies. After execution, Monitor--Inducer generates candidate knowledge from the process, and Curator--Adaptor validates, maintains, retrieves, and uses the knowledge to repair the unfinished plan when needed.

\begin{algorithm}[!htbp]
\caption{Overall MineEvolve Loop}
\label{alg:app_mineevolve_loop}
\begin{algorithmic}[1]
\State \textbf{Input:} task goal $g$, initial state $s_0$, language-model planner $\pi_\theta$, low-level executor $\mathcal{E}$, knowledge store $\mathcal{K}$
\State \textbf{Output:} task outcome and updated knowledge store $\mathcal{K}$

\State Initialize feedback buffer $\mathcal{B}\leftarrow \emptyset$
\State Set current state $s\leftarrow s_0$
\State Curator retrieves skills and remedies from $\mathcal{K}$ relevant to $(g,s)$
\State $\pi_\theta$ generates an initial subgoal plan $\mathbf{z}_{1:N}$ using the goal, state, and retrieved knowledge

\For{$i=1$ to $N$}
    \State Executor $\mathcal{E}$ executes the current subgoal $z_i$ and returns trace $\tau_i$
    \State Monitor--Inducer generates candidate skills or remedies from $\tau_i$ and $\mathcal{B}$
    \State Curator validates, merges, filters, and updates $\mathcal{K}$ with candidate knowledge
    \If{repeated failure or stagnation is detected}
        \State Curator retrieves skills and remedies related to the current failure context
        \State Adaptor freezes the completed prefix and repairs the unfinished plan using retrieved knowledge and active remedies
    \EndIf
    \State Update current state $s$
    \If{task goal $g$ is completed}
        \State \Return success, $\mathcal{K}$
    \EndIf
\EndFor

\State \Return task outcome, $\mathcal{K}$
\end{algorithmic}
\end{algorithm}

\newpage
\subsection{Monitor--Inducer: Feedback-Conditioned Skill and Remedy Generation}
\label{algo2}

Monitor--Inducer corresponds to the feedback-conditioned generation process in the main text. Monitor first compresses a subgoal execution trace into typed execution feedback,
\[
e_i=(z_i,y_i,\Delta s_i,\Delta v_i,f_i,p_i,\ell_i),
\]
where $z_i$ is the current subgoal, $y_i$ is the success label, $\Delta s_i$ and $\Delta v_i$ are the state and inventory changes, $f_i$ is the failure type, $p_i$ is the execution progress, and $\ell_i$ is the stagnation flag. Inducer then generates a skill from successful feedback and a remedy from failed or stagnant segments.

\begin{algorithm}[!htbp]
\caption{Monitor--Inducer: Feedback-conditioned generation of skills and remedies.}
\label{alg:app_monitor_inducer}
\begin{algorithmic}[1]
\State \textbf{Input:} current subgoal $z_i$, execution trace $\tau_i$, feedback buffer $\mathcal{B}$, recent window size $h$
\State \textbf{Output:} updated feedback buffer $\mathcal{B}$ and candidate knowledge set $\mathcal{C}$

\State $\mathcal{C}\leftarrow \emptyset$ \Comment{Initialize candidate skills and remedies}

\Statex \textbf{Stage 1: Monitor extracts typed execution feedback}
\State $(s_i^{\mathrm{pre}}, s_i^{\mathrm{post}}, o_{i,1:T})\leftarrow \Call{ParseTrace}{\tau_i}$
\Comment{Parse pre-state, post-state, and observations}
\State $\Delta s_i\leftarrow \Call{StateDiff}{s_i^{\mathrm{pre}},s_i^{\mathrm{post}}}$
\Comment{Compute environment-state changes}
\State $\Delta v_i\leftarrow \Call{InventoryDiff}{s_i^{\mathrm{pre}},s_i^{\mathrm{post}}}$
\Comment{Compute inventory changes}
\State $y_i\leftarrow \Call{CheckSuccess}{z_i,s_i^{\mathrm{post}}}$
\Comment{Check whether the subgoal succeeds}
\State $f_i\leftarrow \Call{DiagnoseFailure}{z_i,\tau_i,\Delta s_i,\Delta v_i}$
\Comment{Identify failure type}
\State $p_i\leftarrow \Call{ProgressScore}{\tau_i,\Delta s_i,\Delta v_i}$
\Comment{Estimate spatial, inventory, and goal progress}
\State $\ell_i\leftarrow \Call{DetectStagnation}{p_i,y_i}$
\Comment{Detect low-progress or stagnant execution}
\State $e_i\leftarrow (z_i,y_i,\Delta s_i,\Delta v_i,f_i,p_i,\ell_i)$
\Comment{Build typed execution feedback}
\State $\mathcal{B}\leftarrow \mathcal{B}\cup\{e_i\}$
\Comment{Update episode-level feedback buffer}

\Statex \textbf{Stage 2: Inducer identifies feedback segments}
\State $\mathcal{S}^{+}\leftarrow \Call{FindSuccessfulSegments}{\mathcal{B}}$
\Comment{Find consecutive and verifiable successful segments}
\State $\mathcal{S}^{-}\leftarrow \Call{FindFailureSegments}{\mathcal{B},h}$
\Comment{Find recent failed or stagnant segments}

\Statex \textbf{Stage 3: Inducer builds candidate behavioral knowledge}
\ForAll{$\mathcal{B}^{+}\in\mathcal{S}^{+}$}
    \If{\Call{PassTargetCheck}{$\mathcal{B}^{+}$} = True}
        \State $k^{+}\leftarrow \Call{BuildSkill}{\mathcal{B}^{+}}$
        \Comment{Build skill with trigger, preconditions, steps, verification, and effects}
        \State $E(k^{+})\leftarrow \mathcal{B}^{+}$
        \Comment{Record supporting feedback}
        \State $\mathcal{C}\leftarrow \mathcal{C}\cup\{k^{+}\}$
    \EndIf
\EndFor

\ForAll{$\mathcal{B}^{-}\in\mathcal{S}^{-}$}
    \If{\Call{NeedRemedy}{$\mathcal{B}^{-}$} = True}
        \State $k^{-}\leftarrow \Call{BuildRemedy}{\mathcal{B}^{-}}$
        \Comment{Build remedy with trigger, failure, risk, repair, and scope}
        \State $E(k^{-})\leftarrow \mathcal{B}^{-}$
        \Comment{Record failed or stagnant supporting feedback}
        \State $\mathcal{C}\leftarrow \mathcal{C}\cup\{k^{-}\}$
    \EndIf
\EndFor

\State \Return $\mathcal{B},\mathcal{C}$
\end{algorithmic}
\end{algorithm}

Algorithm~\ref{alg:app_monitor_inducer} gives the full Monitor--Inducer procedure. Monitor--Inducer turns raw interaction traces into structured knowledge candidates. Unlike storing complete trajectories or generating free-form reflections, each candidate skill or remedy must include a clear trigger, verification rule, or repair action, which provides a structured basis for Curator validation and retrieval.

\newpage
\subsection{Curator--Adaptor: Knowledge Maintenance and Local Replanning}
\label{algo3}

Algorithm~\ref{alg:app_curator_adaptor} gives the full Curator--Adaptor procedure. Curator--Adaptor prevents MineEvolve from simply placing all historical text into the prompt. Curator first filters out non-executable, overly generic, or conflicting candidates, and then retrieves the most relevant knowledge under a fixed prompt budget. Adaptor uses retrieved skills as positive guidance and remedies as repair knowledge for the unfinished plan.

\begin{algorithm}[!htbp]
\caption{Curator--Adaptor: Knowledge-guided repair of local failures.}
\label{alg:app_curator_adaptor}
\begin{algorithmic}[1]
\State \textbf{Input:} candidate knowledge set $\mathcal{C}$, knowledge store $\mathcal{K}$, task goal $g$, current state $s_i$, current subgoal index $i$, current plan $\mathbf{z}_{1:N}$, feedback buffer $\mathcal{B}$, planner $\pi_\theta$, prompt budget $B$
\State \textbf{Output:} updated knowledge store $\mathcal{K}$ and repaired plan $\mathbf{z}_{1:N}$

\Statex \textbf{Stage 1: Curator validates and maintains candidate knowledge}
\ForAll{$k\in\mathcal{C}$}
    \If{\Call{SchemaCheck}{$k$} = False}
        \State \textbf{continue} \Comment{Discard entries with incomplete fields}
    \EndIf
    \If{\Call{ContextMatch}{$k,s_i$} = False}
        \State \textbf{continue} \Comment{Discard entries not matchable to current state}
    \EndIf
    \If{\Call{ExecutableCheck}{$k$} = False}
        \State \textbf{continue} \Comment{Discard non-executable steps or repair actions}
    \EndIf
    \If{\Call{ConflictCheck}{$k,\mathcal{K}$} = True}
        \State \textbf{continue} \Comment{Discard candidates conflicting with high-confidence knowledge}
    \EndIf
    \If{\Call{SpecificityCheck}{$k$} = False}
        \State \textbf{continue} \Comment{Discard overly generic experience summaries}
    \EndIf
    \State $\mathcal{K}\leftarrow \Call{MergeOrInsert}{\mathcal{K},k}$
    \Comment{Merge similar entries or insert new knowledge}
\EndFor

\Statex \textbf{Stage 2: Curator retrieves planning-relevant knowledge}
\State $c_i\leftarrow (g,s_i,\mathbf{z}_{1:N},\mathcal{B})$
\Comment{Build current planning context}
\State $\mathcal{K}^{R}_i\leftarrow \Call{Retrieve}{\mathcal{K},c_i,B}$
\Comment{Retrieve relevant knowledge under prompt budget $B$}
\State $\mathcal{S}^{R}_i\leftarrow \{k\in\mathcal{K}^{R}_i\mid k.\mathrm{type}=\mathrm{skill}\}$
\Comment{Retrieved reusable skills}
\State $\mathcal{M}^{R}_i\leftarrow \{k\in\mathcal{K}^{R}_i\mid k.\mathrm{type}=\mathrm{remedy}\}$
\Comment{Retrieved remedies}

\Statex \textbf{Stage 3: Adaptor performs knowledge-conditioned local replanning}
\If{\Call{NeedRepair}{$\mathcal{B}$} = True}
    \State $k_{\mathrm{temp}}^{\mathrm{remedy}}\leftarrow \Call{BuildTemporaryRemedy}{\mathcal{B}}$
    \Comment{Build a temporary remedy from recent failure feedback}
    \State $\mathcal{A}_i\leftarrow \mathcal{M}^{R}_i\cup\{k_{\mathrm{temp}}^{\mathrm{remedy}}\}$
    \Comment{Form active remedy set}
    \State $\mathbf{z}_{1:i}\leftarrow \Call{FreezePrefix}{\mathbf{z}_{1:N}, i}$
    \Comment{Keep completed or still valid plan prefix}
    \State $\mathbf{z}'_{i+1:N}\leftarrow
    \pi_\theta(g,s_i,\mathbf{z}_{1:i},\mathcal{S}^{R}_i,\mathcal{A}_i)$
    \Comment{Replan the unfinished subgoals under skills and active remedies}
    \State $\mathbf{z}_{1:N}\leftarrow [\mathbf{z}_{1:i};\mathbf{z}'_{i+1:N}]$
    \Comment{Replace the original suffix with the repaired suffix}
\Else
    \State Keep the current plan $\mathbf{z}_{1:N}$ unchanged
    \Comment{Use retrieved knowledge only in later planning calls}
\EndIf

\State \Return $\mathcal{K},\mathbf{z}_{1:N}$
\end{algorithmic}
\end{algorithm}

For example, if recent feedback shows that the agent repeatedly moves near the same path without inventory changes, Curator maintains a corresponding remedy, and Adaptor asks the planner to avoid the repeated path and generate subgoals such as clearing blocking blocks or taking an alternative route. If the failure is caused by a missing tool or crafting table, Adaptor inserts the corresponding prerequisite into the unfinished plan. Thus, local repair in MineEvolve is a correction of the unfinished plan under both validated external knowledge and current failure feedback.

\newpage
\section{Supplementary Experiments and Analysis}
\label{app:supp_exp}
\subsection{MCU Tech-Tree Task Suite and Metric Definitions}
\label{app:benchmark_details}

We evaluate MineEvolve on a 70-task subset selected from the Minecraft MCU tech-tree benchmark. The task suite is organized according to the Minecraft technology tree and contains seven task groups: Wooden, Stone, Iron, Gold, Redstone, Diamond, and Armor. These groups cover resource collection, tool and weapon crafting, mining, trading, repairing, eating, redstone-component construction, and armor-related tasks. The suite forms a curriculum from short-horizon resource acquisition to long-horizon tasks with deeper prerequisite chains.

Agents operate in a survival-style Minecraft environment. Each episode provides first-person RGB observations and structured state inputs, including the agent's position, health, hunger, GUI status, inventory contents, and observable task progress. The action space covers navigation, camera control, mining, attacking, placing blocks, using items, switching inventory slots, inventory manipulation, and crafting. Unless otherwise specified, each episode starts with an empty inventory, and all task-required resources must be obtained through interaction with the environment.

\begin{table}[!htbp]
\centering
\small
\caption{MCU tech-tree evaluation suite with seven task groups.}
\label{tab:app_task_groups}
\begin{tabularx}{\linewidth}{l c >{\raggedright\arraybackslash}X}
\toprule
\textbf{Group} & \textbf{Task Num.} & \textbf{Representative Task Types} \\
\midrule
Wooden   & 11 & Wooden tools and weapons; wood, planks, sticks, and saplings gathering \\
Stone    & 10 & Stone tools and weapons; stone and cobblestone mining; basic tool upgrades \\
Iron     & 16 & Iron ore mining; smelting; iron tools and weapons; villager trading \\
Gold     & 7  & Gold-related tools; golden foods; repairing golden equipment \\
Redstone & 6  & Redstone components; simple mechanisms; basic circuits and clocks \\
Diamond  & 7  & Diamond mining; diamond tools; repairing and trading for high-tier items \\
Armor    & 13 & Leather and iron armor; washing and repairing armor; helmet-related tasks \\
\bottomrule
\end{tabularx}
\end{table}

\paragraph{Task groups.}
We use Wooden, Stone, and Gold as the easy task groups, since they primarily involve resource acquisition, basic crafting, simple item-level progression, and lower-tier technology-tree objectives in this benchmark subset. We use Iron, Redstone, Diamond, and Armor as the hard task groups, since they involve longer prerequisite chains, more frequent missing-resource failures, tool dependencies, navigation bottlenecks, recovery requirements, and failure propagation across subgoals.

\paragraph{Success rate.}
For a method $m$, task $t$, and evaluation run $r$, let
$s_{m,t,r}\in\{0,1\}$ denote whether the task is successfully completed within the episode horizon. The task-level success rate is defined as
\[
\mathrm{SR}_{m}(t)
=
100 \times \frac{1}{R_t}\sum_{r=1}^{R_t} s_{m,t,r},
\]
where $R_t$ is the number of evaluation runs for task $t$.

For each task group $g$, let $\mathcal{T}_g$ denote the set of tasks in that group. The group-level success rate is
\[
\mathrm{SR}_{m}(g)
=
\frac{1}{|\mathcal{T}_g|}
\sum_{t\in \mathcal{T}_g}
\mathrm{SR}_{m}(t).
\]

\paragraph{Aggregated metrics.}
The main text reports three aggregate metrics: Easy Avg., Hard Avg., and Overall. 
All aggregate metrics are computed as task-count-weighted averages, so that each individual task contributes equally to the corresponding score. 
This avoids assigning the same weight to task groups with different numbers of tasks.

Let \(n_g = |\mathcal{T}_g|\) denote the number of tasks in group \(g\). 
For the 70-task MCU subset, the group sizes are:
\[
\begin{aligned}
&n_{\mathrm{Wooden}}=11,\quad
n_{\mathrm{Stone}}=10,\quad
n_{\mathrm{Iron}}=16,\quad
n_{\mathrm{Gold}}=7,\\
&n_{\mathrm{Redstone}}=6,\quad
n_{\mathrm{Diamond}}=7,\quad
n_{\mathrm{Armor}}=13.
\end{aligned}
\]
Easy Avg. is the task-count-weighted average over the easy task groups, namely Wooden, Stone, and Gold:
\[
\mathrm{EasyAvg}_{m}
=
\frac{
11\,\mathrm{SR}_{m}(\mathrm{Wooden})
+
10\,\mathrm{SR}_{m}(\mathrm{Stone})
+
7\,\mathrm{SR}_{m}(\mathrm{Gold})
}{28}.
\]

Hard Avg. is the task-count-weighted average over the hard task groups, namely Iron, Redstone, Diamond, and Armor:
\[
\mathrm{HardAvg}_{m}
=
\frac{
16\,\mathrm{SR}_{m}(\mathrm{Iron})
+
6\,\mathrm{SR}_{m}(\mathrm{Redstone})
+
7\,\mathrm{SR}_{m}(\mathrm{Diamond})
+
13\,\mathrm{SR}_{m}(\mathrm{Armor})
}{42}.
\]

Overall is computed using the same task-count-weighted convention over all seven task groups, i.e., as the average task-level success rate over the full 70-task suite.

\paragraph{Protocol control.}
All compared methods are evaluated on the same task set, task--seed split, episode horizon, success criterion, and success-rate metric. For STEVE-1-based methods, we additionally fix the low-level execution policy, observation/action interface, action primitives, structured state fields, retrieval top-$K$, memory-token budget, and evaluation-time LLM-call budget. This ensures that performance differences primarily reflect differences in planning, execution-feedback utilization, knowledge maintenance, and local repair, rather than differences in task selection or evaluation protocol.
\subsection{Full Task-Group Results}
\label{app:full_results}

Table~\ref{tab:app_full_results} provides the full success rates across all seven MCU task groups. This table complements the compact results in Table~\ref{tab:main_results_panel} by showing how the gains are distributed across different stages of the Minecraft tech tree. We additionally report the STEVE-1 low-level-only baseline, which highlights the difficulty of solving long-horizon dependency tasks without high-level planning and recovery.

\begin{table*}[!htbp]
\centering
\small
\caption{Full results on the MCU task suite. Values denote success rate SR (\%). All system-level methods are evaluated under the same task set, evaluation metric, and interaction horizon. Wo., St., Ir., Go., Re., Di., and Ar. denote Wooden, Stone, Iron, Gold, Redstone, Diamond, and Armor, respectively. Overall is task-weighted over all 70 tasks.}
\label{tab:app_full_results}
\resizebox{\textwidth}{!}{
\begin{tabular}{llcccccccc}
\toprule
Planner backbone & Method & Wo. & St. & Ir. & Go. & Re. & Di. & Ar. & Overall \\
\midrule
\multirow{5}{*}{Qwen3.5-Flash}
& DEPS & 82.91 & 68.37 & 16.48 & 2.79 & 5.84 & 2.11 & 7.25 & 28.90 \\
& JARVIS-1 & 92.37 & 88.14 & 34.76 & \textbf{10.83} & 23.41 & 8.39 & 14.92 & 41.75 \\
& Optimus-1 & 98.30 & \textbf{92.75} & 45.24 & 8.07 & 28.17 & 8.21 & 14.80 & 45.83 \\
& Optimus-2 (GOAP) & \textbf{98.60} & 92.45 & 51.30 & 8.70 & 28.10 & 12.40 & 20.95 & 48.84 \\
& MineEvolve & 98.09 & 92.61 & \textbf{52.43} & 10.11 & \textbf{30.77} & \textbf{14.58} & \textbf{23.47} & \textbf{50.09} \\
\midrule
\multirow{5}{*}{Qwen3.5-Plus}
& DEPS & 84.05 & 69.34 & 17.39 & 3.47 & 8.94 & 2.43 & 4.36 & 29.25 \\
& JARVIS-1 & 93.08 & 88.73 & 35.57 & 11.12 & 26.18 & 9.03 & 15.58 & 42.59 \\
& Optimus-1 & 97.91 & \textbf{93.60} & 46.06 & 7.05 & \textbf{32.03} & 11.42 & 19.09 & 47.42 \\
& Optimus-2 (GOAP) & \textbf{99.05} & 93.20 & 53.20 & 9.20 & 28.70 & 13.20 & 21.60 & 49.75 \\
& MineEvolve & 98.71 & 93.48 & \textbf{55.83} & \textbf{13.79} & 31.24 & \textbf{17.06} & \textbf{27.63} & \textbf{52.52} \\
\midrule
\multirow{5}{*}{GLM-4.7}
& DEPS & 82.37 & 66.91 & 15.92 & 1.36 & 7.82 & 2.18 & 4.12 & 27.93 \\
& JARVIS-1 & 91.79 & 86.87 & 33.94 & \textbf{8.87} & 21.82 & 8.08 & 14.19 & 40.79 \\
& Optimus-1 & 96.63 & 91.40 & 44.13 & 7.79 & 23.06 & 7.95 & 18.04 & 45.23 \\
& Optimus-2 (GOAP) & \textbf{97.90} & \textbf{91.65} & 50.35 & 8.35 & 26.85 & \textbf{12.05} & 20.10 & 48.06 \\
& MineEvolve & 97.36 & 91.18 & \textbf{50.55} & 8.63 & \textbf{29.08} & 10.92 & \textbf{22.07} & \textbf{48.43} \\
\midrule
\multirow{5}{*}{Gemini-3-Flash}
& DEPS & 84.25 & 69.86 & 17.81 & 4.07 & 9.61 & 2.76 & 4.66 & 29.66 \\
& JARVIS-1 & 92.99 & 88.58 & 35.21 & 9.79 & 22.96 & 8.92 & 15.52 & 42.04 \\
& Optimus-1 & 98.70 & 91.76 & 45.92 & 6.85 & 24.48 & 13.06 & 18.98 & 46.73 \\
& Optimus-2 (GOAP) & \textbf{99.10} & \textbf{93.55} & 54.20 & 10.05 & 30.25 & \textbf{13.80} & 22.35 & 50.45 \\
& MineEvolve & 98.63 & 93.37 & \textbf{55.28} & \textbf{13.52} & \textbf{33.73} & 12.78 & \textbf{27.18} & \textbf{52.04} \\
\midrule
\multirow{5}{*}{GPT-5.5}
& DEPS & 86.42 & 72.38 & 20.31 & 6.16 & 11.84 & 3.68 & 6.07 & 31.69 \\
& JARVIS-1 & 94.18 & 90.31 & 38.44 & 12.61 & 25.96 & 10.34 & 16.14 & 44.01 \\
& Optimus-1 & 98.62 & \textbf{94.82} & 49.28 & 10.71 & 31.47 & 15.31 & 17.86 & 48.92 \\
& Optimus-2 (GOAP) & \textbf{99.22} & 94.60 & 56.15 & 11.35 & 32.10 & 14.70 & 23.50 & 51.66 \\
& MineEvolve & 99.16 & 94.77 & \textbf{60.12} & \textbf{16.70} & \textbf{36.47} & \textbf{18.91} & \textbf{28.98} & \textbf{54.93} \\
\midrule
STEVE-1 & low-level only & 25.61 & 18.72 & 2.47 & 0.00 & 0.00 & 0.00 & 0.00 & 7.26 \\
\bottomrule
\end{tabular}
}
\end{table*}
\subsection{Positioning with Recent Systems}
\label{app:recent_positioning}

Table~\ref{tab:app_recent_positioning} summarizes the relationship between MineEvolve and recent Minecraft agents from the perspectives of experimental protocol and mechanism design. The purpose of this table is to clarify the research focus of different systems, rather than to provide an absolute ranking under mismatched evaluation protocols.

\begin{table}[!htbp]
\centering
\small
\caption{Mechanistic positioning of recent Minecraft agents.}
\label{tab:app_recent_positioning}
\resizebox{\linewidth}{!}{
\begin{tabular}{lccccc}
\toprule
Method & Low-level policy & Use of failure feedback & Typed feedback & Continual knowledge update & Same-policy comparison \\
\midrule
DEPS & STEVE-1 & Feedback interpretation and subgoal selection & No & No & Yes \\
JARVIS-1 & STEVE-1 & Weak; mainly successful-case retrieval & No & No & Yes \\
Optimus-1 & STEVE-1 & Retrieved as part of the experience pool & No & Partial & Yes \\
Optimus-2 & GOAP & Not a core mechanism & No & No & No \\
MineEvolve & STEVE-1 & Remedy generation & Yes & Yes & Yes \\
\bottomrule
\end{tabular}
}
\end{table}

\subsection{Knowledge Accumulation and Runtime Overhead}
\label{app:accumulation_overhead}

Table~\ref{tab:app_accumulation_by_group} reports the group-wise results of the knowledge accumulation experiment on the four difficult task groups. At each checkpoint, we freeze the external knowledge base before evaluation. Hard Avg. denotes the average success rate over Iron, Redstone, Diamond, and Armor. 
\begin{table*}[!htbp]
\centering
\tiny
\caption{Group-wise results for the knowledge accumulation curves. Values denote SR (\%) on held-out hard tasks. Hard Avg. is the task-count-weighted average over Iron, Redstone, Diamond, and Armor, using group sizes 16, 6, 7, and 13, respectively. At each checkpoint, the external knowledge base is frozen before evaluation.}
\label{tab:app_accumulation_by_group}
\resizebox{\textwidth}{!}{
\begin{tabular}{llccccc}
\toprule
Method & Task group & 0 eps. & 50 eps. & 100 eps. & 200 eps. & 400 eps. \\
\midrule
\multirow{5}{*}{Static Store}
& Iron     & 39.8 & 40.8 & 41.9 & 42.9 & 43.6 \\
& Redstone & 24.2 & 24.9 & 25.6 & 26.2 & 26.5 \\
& Diamond  & 9.0  & 9.5  & 9.9  & 10.4 & 10.9 \\
& Armor    & 16.2 & 17.2 & 18.2 & 19.3 & 19.4 \\
& Hard Avg. & 25.13 & 26.01 & 26.90 & 27.79 & 28.22 \\
\midrule
\multirow{5}{*}{Text Reflection}
& Iron     & 39.8 & 42.3 & 44.1 & 44.9 & 45.4 \\
& Redstone & 24.2 & 25.6 & 26.8 & 27.4 & 27.9 \\
& Diamond  & 9.0  & 10.2 & 11.0 & 11.5 & 12.0 \\
& Armor    & 16.2 & 18.7 & 19.3 & 20.2 & 20.3 \\
& Hard Avg. & 25.13 & 27.26 & 28.44 & 29.19 & 29.56 \\
\midrule
\multirow{5}{*}{MineEvolve}
& Iron     & 39.8 & 45.4 & 48.2 & 51.3 & 53.4 \\
& Redstone & 24.2 & 27.2 & 28.6 & 30.0 & 31.0 \\
& Diamond  & 9.0  & 11.2 & 12.7 & 14.4 & 15.8 \\
& Armor    & 16.2 & 21.4 & 22.5 & 23.5 & 24.6 \\
& Hard Avg. & 25.13 & \textbf{29.67} & \textbf{31.53} & \textbf{33.50} & \textbf{35.02} \\
\bottomrule
\end{tabular}
}
\end{table*}

Since MineEvolve relies on an external knowledge base, it is important to examine whether knowledge growth introduces excessive retrieval or prompting overhead. Table~\ref{tab:app_overhead} reports the number of skills, the number of remedies, the average number of retrieved tokens, retrieval latency, and the number of evaluation-time LLM calls at different accumulation checkpoints. This analysis verifies whether the budget control of the Curator prevents unbounded context growth.

\begin{table}[!htbp]
\centering
\tiny
\caption{Knowledge-base size and runtime overhead. Values are averaged over held-out hard-task episodes.}
\label{tab:app_overhead}
\resizebox{\linewidth}{!}{
\begin{tabular}{lccccc}
\toprule
Checkpoint & \#Skills & \#Remedies & Retrieved tokens & Retrieval latency & Eval. LLM calls \\
\midrule
0 eps. & 0 & 0 & 0 & 0.00 s & 7.82 \\
50 eps. & 31 & 44 & 628 & 0.10 s & 7.96 \\
100 eps. & 56 & 78 & 846 & 0.14 s & 8.07 \\
200 eps. & 94 & 128 & 1087 & 0.19 s & 8.18 \\
400 eps. & 143 & 191 & 1216 & 0.25 s & 8.31 \\
\bottomrule
\end{tabular}
}
\end{table}

\subsection{Curriculum-style Knowledge Construction on Diamond Tasks}
\label{app:diamond_curriculum}

To further analyze how the source of external behavioral knowledge affects difficult tasks, we conduct a curriculum-style knowledge construction experiment on the Diamond task group. This experiment differs from the unified MCU evaluation protocol used in the main experiments; instead, it is designed to isolate how different knowledge initialization and update strategies influence continual learning efficiency on Diamond tasks. All settings share the same planner backbone, STEVE-1 low-level execution policy, environment interface, interaction horizon, retrieval budget, and evaluation-time LLM-call budget. The only difference lies in how the external knowledge bases $K_{\mathrm{skill}}$ and $K_{\mathrm{remedy}}$ are initialized and updated.

Table~\ref{tab:diamond_curriculum_protocol} summarizes the four knowledge construction strategies. Cold Start keeps the knowledge base empty throughout and serves as a lower bound without cross-episode behavioral knowledge. Diamond-only Self-learning generates skills and remedies only from Diamond tasks. Curriculum Pretrain$\rightarrow$Freeze first builds the knowledge base on lower-tier tasks from Wooden to Redstone, and then freezes this knowledge when transferring to Diamond tasks. Mixed Sampling (1:1) mixes lower-tier tasks and Diamond tasks at a 1:1 ratio and maintains a unified $K_{\mathrm{skill}}/K_{\mathrm{remedy}}$ online.

\begin{table}[!htbp]
\centering
\small
\caption{Curriculum-style knowledge construction strategies for Diamond tasks. All settings share the same planner, low-level execution policy, interaction horizon, and retrieval budget.}
\label{tab:diamond_curriculum_protocol}
\resizebox{\linewidth}{!}{
\begin{tabular}{lccc}
\toprule
Strategy & Knowledge initialization & Training-stage update & Purpose \\
\midrule
Cold Start (Empty KB)
& Empty $K_{\mathrm{skill}}/K_{\mathrm{remedy}}$
& No cross-episode knowledge writing
& Estimate the lower bound without external behavioral knowledge \\
Diamond-only Self-learning
& Empty $K_{\mathrm{skill}}/K_{\mathrm{remedy}}$
& Write Diamond-only experience
& Test whether self-learning only on the target hard tasks is sufficient \\
Curriculum Pretrain$\rightarrow$Freeze
& Knowledge pretrained on lower-tier tasks
& Freeze lower-tier knowledge during the Diamond stage
& Test whether lower-tier knowledge transfers to Diamond tasks \\
Mixed Sampling (1:1)
& Unified online knowledge base
& Jointly update with lower-tier and Diamond experience
& Test whether mixed updates yield stronger continual improvement \\
\bottomrule
\end{tabular}
}
\end{table}

The training budget is controlled by the total number of episodes, and evaluation is performed at five checkpoints: 20\%, 40\%, 60\%, 80\%, and 100\%. For Diamond-only Self-learning and Mixed Sampling (1:1), each checkpoint corresponds to the completed online training progress under the respective strategy. For Curriculum Pretrain$\rightarrow$Freeze, each checkpoint corresponds to the construction progress of the lower-tier pretrained knowledge base. At each checkpoint, we freeze the current knowledge base before evaluating on Diamond tasks. Each evaluation is repeated across multiple world seeds, and the success rate is averaged over the seven Diamond tasks. Table~\ref{tab:diamond_curriculum_sr} reports the success rates on Diamond tasks. This experiment is intended to analyze the effect of knowledge construction pathways on Diamond tasks and does not replace the unified MCU evaluation results reported in the main table.
\begin{table}[!htbp]
\centering
\small
\caption{Success rates of different curriculum-style knowledge construction strategies on Diamond tasks. Values denote SR (\%), averaged over seven Diamond tasks and repeated across multiple world seeds.}
\label{tab:diamond_curriculum_sr}
\resizebox{\linewidth}{!}{
\begin{tabular}{lccccc}
\toprule
Strategy & SR@20\% & SR@40\% & SR@60\% & SR@80\% & SR@100\% \\
\midrule
Cold Start (Empty KB)
& 2.6 & 2.8 & 2.7 & 2.8 & 2.9 \\
Diamond-only Self-learning
& 3.0 & 6.1 & 9.6 & 12.2 & 13.9 \\
Curriculum Pretrain$\rightarrow$Freeze
& 5.7 & 8.9 & 11.8 & 14.3 & 16.4 \\
Mixed Sampling (1:1)
& \textbf{6.2} & \textbf{9.8} & \textbf{13.1} & \textbf{15.8} & \textbf{17.6} \\
\bottomrule
\end{tabular}
}
\end{table}
As shown in Table~\ref{tab:diamond_curriculum_sr}, different knowledge construction strategies lead to clear performance differences on Diamond tasks. Cold Start (Empty KB) remains below 3\% across all checkpoints, with the success rate only changing slightly from 2.6\% to 2.9\%. This suggests that, without cross-episode skills and remedies, the agent struggles to achieve sustained improvement on long-dependency tasks such as Diamond. Diamond-only Self-learning improves steadily as training progresses, increasing from 3.0\% to 13.9\%. This indicates that task-specific knowledge can be gradually induced from Diamond trajectories alone. However, its low early-stage success rate also suggests that accumulating experience from scratch on high-tier tasks is sample-inefficient.

In contrast, Curriculum Pretrain$\rightarrow$Freeze consistently outperforms Diamond-only Self-learning at all checkpoints and reaches a final success rate of 16.4\%. This result shows that knowledge induced from lower-tier tasks, including tool crafting, resource collection, smelting, crafting prerequisites, and common failure-recovery patterns, can transfer to Diamond tasks. Although this strategy does not continue updating the lower-tier knowledge base during the Diamond stage, its performance on Diamond tasks improves as the lower-tier pretrained knowledge base becomes more complete across checkpoints.

Mixed Sampling (1:1) achieves the highest success rate at all checkpoints, improving from 6.2\% to 17.6\%. Compared with Curriculum Pretrain$\rightarrow$Freeze, Mixed Sampling not only preserves transferable foundational knowledge from lower-tier tasks, but also continuously incorporates Diamond-specific skills and remedies during training. As a result, it maintains a stable advantage in the middle and later stages, suggesting that for high-tier tasks, freezing lower-tier knowledge provides a strong initialization, while continually mixing lower-tier and target-task experience is more effective for capturing high-tier-specific failure modes and repair actions.

To more directly compare the overall learning efficiency of different strategies, Table~\ref{tab:diamond_curriculum_gain} summarizes the average success rate over the five checkpoints and the final success rate. Mixed Sampling (1:1) achieves the highest checkpoint-averaged success rate, indicating that its advantage is not limited to the final stage but remains stable throughout training. Curriculum Pretrain$\rightarrow$Freeze ranks second in checkpoint average, further validating the effectiveness of lower-tier knowledge transfer. Diamond-only Self-learning eventually achieves a substantial improvement, but its overall learning efficiency is lower. Cold Start remains close to the performance floor throughout.

\begin{table}[!htbp]
\centering
\tiny
\caption{Summary of gains from curriculum-style knowledge construction on Diamond tasks. Checkpoint Avg. denotes the average success rate over the five checkpoints. Final gain is computed relative to Diamond-only Self-learning.}
\label{tab:diamond_curriculum_gain}
\resizebox{\linewidth}{!}{
\begin{tabular}{lccc}
\toprule
Strategy & Checkpoint Avg. & Final SR & Final gain vs. Diamond-only \\
\midrule
Cold Start (Empty KB) & 2.76 & 2.9 & -11.0 \\
Diamond-only Self-learning & 8.96 & 13.9 & 0.0 \\
Curriculum Pretrain$\rightarrow$Freeze & 11.42 & 16.4 & +2.5 \\
Mixed Sampling (1:1) & \textbf{12.50} & \textbf{17.6} & \textbf{+3.7} \\
\bottomrule
\end{tabular}
}
\end{table}

Overall, this experiment shows that performance on Diamond tasks depends not only on the number of training episodes, but also on the pathway through which the knowledge base is constructed. External behavioral knowledge acquired from lower-tier tasks substantially improves cold-start performance on high-tier tasks, while continual mixed updates further improve the final success rate. This finding is consistent with the design goal of MineEvolve: the agent's capability growth does not come from updating the parameters of the language model, but from continually accumulating, validating, and reusing external skills and remedies.
\subsection{Qualitative Case Studies}
\label{app:case_study}

In this section, we present two qualitative case studies to illustrate how MineEvolve transforms successful and failed execution experiences into external behavioral knowledge that can affect subsequent planning. The first case demonstrates failure-conditioned local replanning: when the agent becomes stuck near a terrain obstacle, the system converts the failure into a remedy that can be immediately applied within the current episode, and the Adaptor repairs the remaining plan accordingly. The second case illustrates cross-episode knowledge accumulation: a failed crafting attempt is distilled into a reusable remedy, while a subsequent successful execution is stored in the external knowledge base as a skill.

\begin{figure}[htbp]
    \centering
    \includegraphics[width=0.98\linewidth]{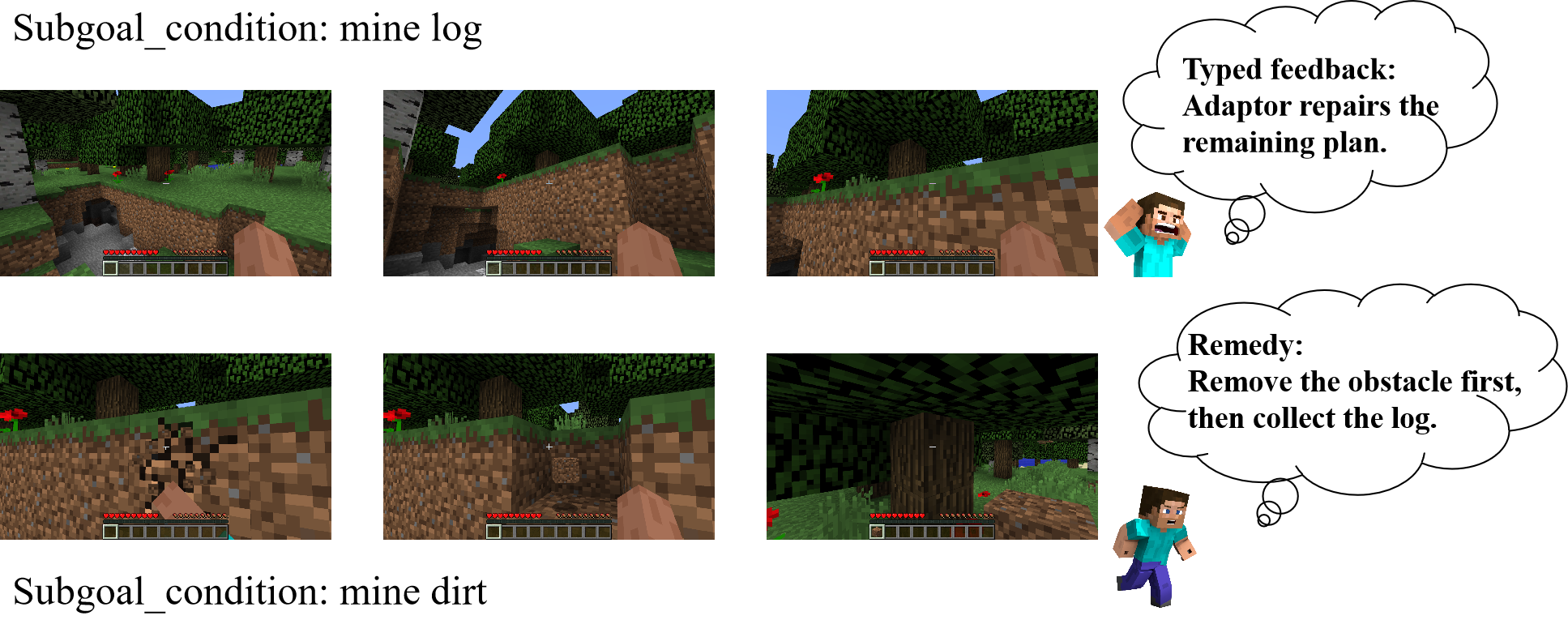}
    \caption{Failure-conditioned local replanning. When the agent repeatedly moves near a terrain obstacle without making effective progress and without any inventory change, the Monitor identifies the behavior as a low-progress navigation failure. The resulting remedy prompts the Adaptor to repair the remaining plan by inserting a corrective subgoal: first break the dirt block blocking the path, and then continue collecting the target wood.}
    \label{fig:case-replan}
\end{figure}

Figure~\ref{fig:case-replan} illustrates how typed execution feedback supports local plan repair. The agent initially attempts to approach the target wood, but repeatedly moves near a terrain boundary with only limited change in position, while the inventory shows no increase in the target resource. The Monitor compresses this execution trace into typed feedback, including the failed subgoal, low-progress signals, no inventory change, and a navigation-related failure type. Because the failure is not a one-off mistake but a repeated stagnation pattern, the Inducer further generates an active remedy that specifies both triggering conditions and a repair action: instead of repeatedly following the same approach path, the agent should first remove the blocking terrain or choose an alternative route.

The Curator then retrieves or maintains this remedy according to the current spatial context, and the Adaptor preserves the valid plan prefix while repairing only the unfinished downstream subgoals. As a result, the original plan no longer continues to approach the wood directly, but is revised to first break the dirt block blocking the path and then resume the subgoal of collecting the target wood. This case highlights that a remedy in MineEvolve is not merely a natural-language explanation of failure, but an executable constraint with explicit triggers and repair actions that can directly modify the remaining plan within the current episode.

\begin{figure}[htbp]
    \centering
    \includegraphics[width=0.98\linewidth]{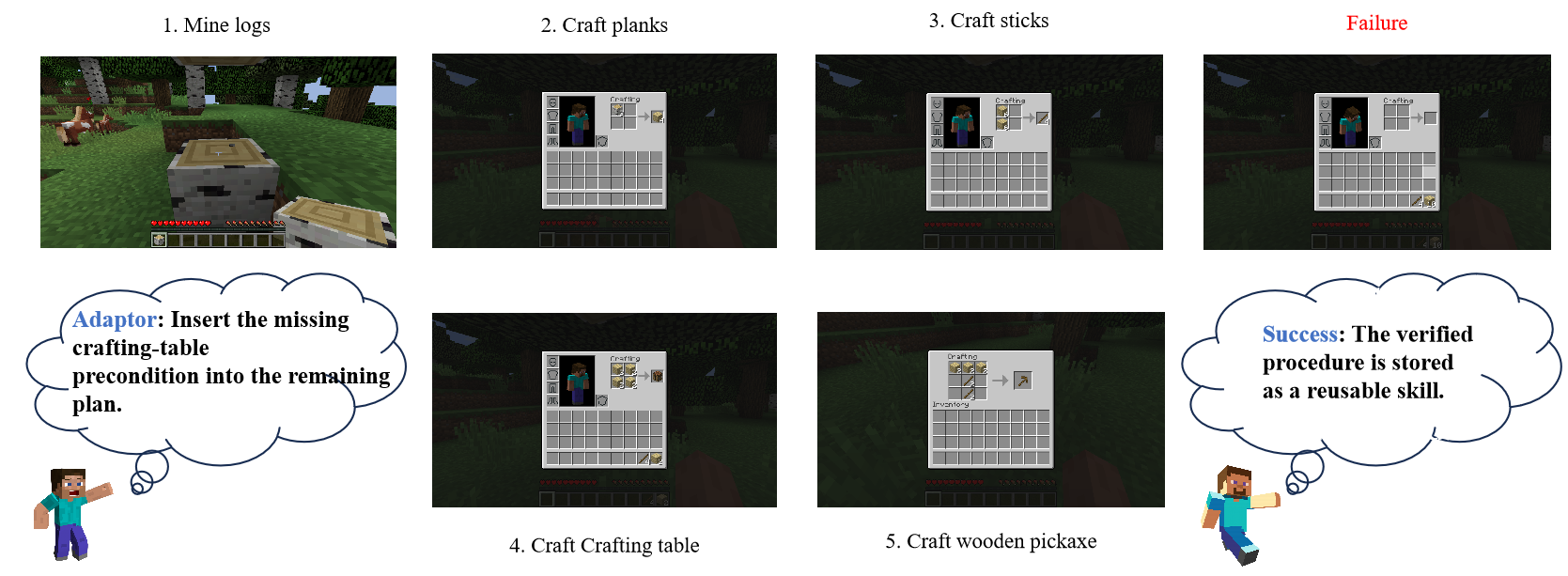}
    \caption{Feedback-conditioned generation of skills and remedies. A failed crafting attempt is converted into a remedy, indicating that recipes requiring a larger crafting grid must first obtain and place a crafting table. Once the corrected execution succeeds, the successful process is further stored as a reusable skill.}
    \label{fig:case-distill}
\end{figure}

Figure~\ref{fig:case-distill} shows how MineEvolve accumulates external behavioral knowledge across episodes. The agent initially attempts to craft the target tool directly in the inventory crafting interface, but the recipe requires the larger crafting grid provided by a crafting table, causing the execution to fail. The Monitor records this failure through fields such as GUI state, inventory changes, and the absence of the target item, and summarizes it as typed execution feedback. Based on this feedback, the Inducer generates a remedy: when a crafting failure is caused by the missing prerequisite of a crafting table, the agent should first craft and place a crafting table before retrying the recipe.

Before writing the candidate remedy into the knowledge base, the Curator validates it by checking whether the triggering conditions can be matched by the current state fields, whether the repair action is executable by the low-level policy, and whether the knowledge is sufficiently specific and does not conflict with existing high-confidence knowledge. Once validated, the remedy is stored in the external knowledge base and can be retrieved in future similar tasks. Later, when the agent again plans a crafting task that requires a crafting table, the Adaptor inserts the prerequisite step of ``craft and place a crafting table'' into the remaining plan based on the retrieved remedy. If the corrected execution succeeds, the Inducer further abstracts this successful process into a skill, recording its trigger conditions, prerequisites, steps, execution effects, and verification rules. This case illustrates that MineEvolve leverages both failure and success: failures generate remedies to avoid repeated mistakes, while successful executions generate skills to reuse validated procedural knowledge.

\newpage
\section{Prompt Design and Tool-Calling Details}
\label{app:prompt_details}
This section provides the prompt templates used by MineEvolve. The prompts are aligned with the four-stage procedure in the main method: \textit{Monitor} extracts typed execution feedback, \textit{Inducer} converts feedback into candidate skills and remedies, \textit{Curator} validates and retrieves relevant knowledge, and \textit{Adaptor} repairs only the unfinished part of the plan under retrieved skills and active remedies. All prompts use a chat-style interface with explicit \texttt{system} and \texttt{user} roles. To ensure downstream parsing and validation, the LLM is required to output strictly structured JSON objects.

\paragraph{Planner: system prompt and JSON schema.}
The planner generates a short sequence of high-level subgoals. Each subgoal must be executable by the low-level executor and must include explicit completion checks. The planner is also instructed to use retrieved skills as positive procedural guidance and remedies as constraints or repair knowledge.

\noindent\framebox[\linewidth][l]{%
	\parbox{0.97\linewidth}{\small
		\textbf{Planner system prompt.}\\[0.5ex]
		\texttt{You are the high-level planner of MineEvolve, a Minecraft embodied agent.}\\
		\texttt{Generate a short executable plan for the current task. Use retrieved skills as}\\
		\texttt{positive guidance and active remedies as constraints for avoiding repeated failures.}\\[0.5ex]
		\texttt{Output JSON only. Use the following schema:}\\
		\texttt{\{}\\
		\quad\texttt{"plan\_id": "p\_xxxx",}\\
		\quad\texttt{"subgoals": [}\\
		\quad\quad\texttt{\{}\\
		\quad\quad\quad\texttt{"subgoal\_id": "sg\_001",}\\
		\quad\quad\quad\texttt{"condition": "short executable action",}\\
		\quad\quad\quad\texttt{"task\_kind": "mine|craft|smelt|use|combat|move|wait",}\\
		\quad\quad\quad\texttt{"executor\_hint": "stevei|mcu\_craft|mcu\_smelt|wait",}\\
		\quad\quad\quad\texttt{"mode": "move|stay",}\\
		\quad\quad\quad\texttt{"timeout\_s": 60,}\\
		\quad\quad\quad\texttt{"checks": [\{"type":"inv\_ge","item":"oak\_log","n":1\}],}\\
		\quad\quad\quad\texttt{"rationale": "one short reason"}\\
		\quad\quad\texttt{\}}\\
		\quad\texttt{],}\\
		\quad\texttt{"global\_constraints": []}\\
		\texttt{\}}\\[0.5ex]
		\texttt{Planning rules:}\\
		\texttt{1) Keep only necessary prerequisites for the target task;}\\
		\texttt{2) Reuse retrieved skills when their preconditions match the current state;}\\
		\texttt{3) Obey active remedies and avoid repeating known failure patterns;}\\
		\texttt{4) Every subgoal must have a verifiable completion check;}\\
		\texttt{5) Do not output explanations outside JSON.}
	}%
}

\newpage
\paragraph{Planner: user prompt with retrieved knowledge.}
The user prompt injects the current task, environment state, current plan prefix when available, retrieved skills, and active remedies. Unlike a generic memory capsule, the injected knowledge is separated into skills and remedies maintained by the Curator.

\noindent\framebox[\linewidth][l]{%
	\parbox{0.97\linewidth}{\small
		\textbf{Planner user prompt.}\\[0.5ex]
		\texttt{Task goal: craft an iron pickaxe}\\[0.25ex]
		\texttt{Current state:}\\
		\texttt{\ \ inventory=\{"iron\_ingot":3,"stick":2\}, coords=[100,64,200], HP=20, hunger=20}\\[0.25ex]
		\texttt{Current plan prefix:}\\
		\texttt{\ \ completed=[]}\\[0.25ex]
		\texttt{Retrieved skills from Curator:}\\
		\texttt{\ \ - id: skill\_craft\_tool\_at\_table}\\
		\texttt{\ \ \ \ trigger: target requires 3x3 crafting grid}\\
		\texttt{\ \ \ \ preconditions: ["have required ingredients", "crafting table available"]}\\
		\texttt{\ \ \ \ steps: ["place crafting table", "open crafting table", "craft target item"]}\\
		\texttt{\ \ \ \ verification: inventory contains target item}\\[0.25ex]
		\texttt{Active remedies from Curator/Inducer:}\\
		\texttt{\ \ - id: remedy\_missing\_crafting\_table}\\
		\texttt{\ \ \ \ trigger: crafting fails because no 3x3 grid is available}\\
		\texttt{\ \ \ \ repair: insert obtain-or-place-crafting-table before retrying recipe}\\
		\texttt{\ \ \ \ scope: unfinished plan suffix}\\[0.5ex]
		\texttt{Generate the shortest executable plan that reaches the task goal.}
	}%
}

\newpage
\paragraph{Inducer: skill generation from successful feedback.}
When Monitor produces a sequence of successful and verifiable feedback records, Inducer asks the LLM to abstract them into a reusable skill. The output follows the unified knowledge-entry format used by MineEvolve.

\noindent\framebox[\linewidth][l]{%
	\parbox{0.97\linewidth}{\small
		\textbf{Skill induction prompt.}\\[0.5ex]
		\texttt{System:}\\
		\texttt{You are the Inducer module of MineEvolve. Convert successful typed execution}\\
		\texttt{feedback into one reusable skill. The skill must be specific, executable, and verifiable.}\\
		\texttt{Output JSON only.}\\[0.5ex]
		\texttt{User:}\\
		\texttt{Task goal: craft sticks}\\
		\texttt{Successful feedback segment:}\\
		\texttt{\ \ - e\_001: z="collect oak log", y=1, delta\_inv=\{"oak\_log":4\}, f="NONE", p=1.0, stagnant=false}\\
		\texttt{\ \ - e\_002: z="craft planks", y=1, delta\_inv=\{"oak\_planks":16\}, f="NONE", p=1.0, stagnant=false}\\
		\texttt{\ \ - e\_003: z="craft sticks", y=1, delta\_inv=\{"stick":4\}, f="NONE", p=1.0, stagnant=false}\\[0.5ex]
		\texttt{Output schema:}\\
		\texttt{\{}\\
		\quad\texttt{"type": "skill",}\\
		\quad\texttt{"id": "skill\_xxxx",}\\
		\quad\texttt{"trigger\_context": ["..."],}\\
		\quad\texttt{"preconditions": ["..."],}\\
		\quad\texttt{"steps": ["..."],}\\
		\quad\texttt{"verification": \{"type":"inv\_ge","item":"stick","n":4\},}\\
		\quad\texttt{"observed\_effects": ["..."],}\\
		\quad\texttt{"supporting\_feedback": ["e\_001","e\_002","e\_003"],}\\
		\quad\texttt{"confidence": 0.8}\\
		\texttt{\}}
	}%
}

\newpage
\paragraph{Inducer: remedy generation from failed or stagnant feedback.}
When Monitor detects repeated failures or low-progress execution, Inducer generates a remedy rather than a free-form reflection. A remedy must specify when it applies, what risk pattern it captures, and what repair action should be inserted into the unfinished plan.

\noindent\framebox[\linewidth][l]{%
	\parbox{0.97\linewidth}{\small
		\textbf{Remedy induction prompt.}\\[0.5ex]
		\texttt{System:}\\
		\texttt{You are the Inducer module of MineEvolve. Convert failed or stagnant typed execution}\\
		\texttt{feedback into one executable remedy. The remedy must include a trigger, failure type,}\\
		\texttt{risk pattern, repair action, and repair scope. Do not output generic advice such as}\\
		\texttt{"try again" or "be careful". Output JSON only.}\\[0.5ex]
		\texttt{User:}\\
		\texttt{Current subgoal: collect oak log}\\
		\texttt{Recent feedback window:}\\
		\texttt{\ \ - e\_121: y=0, delta\_inv=\{\}, f="NAV\_STUCK", p=0.05, stagnant=true,}\\
		\texttt{\ \ \ \ coords=[118,64,208], note="repeated movement near same obstacle"}\\
		\texttt{\ \ - e\_122: y=0, delta\_inv=\{\}, f="NAV\_STUCK", p=0.04, stagnant=true,}\\
		\texttt{\ \ \ \ coords=[119,64,208], note="no inventory progress"}\\[0.5ex]
		\texttt{Output schema:}\\
		\texttt{\{}\\
		\quad\texttt{"type": "remedy",}\\
		\quad\texttt{"id": "remedy\_xxxx",}\\
		\quad\texttt{"trigger\_context": ["navigation failure", "low progress", "no inventory gain"],}\\
		\quad\texttt{"failure\_type": "NAV\_STUCK",}\\
		\quad\texttt{"risk\_pattern": "agent repeats movement near obstacle without collecting target item",}\\
		\quad\texttt{"repair\_action": ["clear blocking block or choose alternative route",}\\
		\quad\quad\texttt{"then retry collecting target resource"],}\\
		\quad\texttt{"scope": "unfinished\_plan\_suffix",}\\
		\quad\texttt{"applicability": ["same or similar navigation context", "target item not increased"],}\\
		\quad\texttt{"supporting\_feedback": ["e\_121","e\_122"],}\\
		\quad\texttt{"confidence": 0.8}\\
		\texttt{\}}
	}%
}

\newpage
\paragraph{Curator: validation and retrieval of knowledge entries.}
Curator prevents vague, conflicting, or non-executable knowledge from entering the external knowledge store. It also retrieves a compact set of relevant skills and remedies under the prompt budget.

\noindent\framebox[\linewidth][l]{%
	\parbox{0.97\linewidth}{\small
		\textbf{Curator prompt.}\\[0.5ex]
		\texttt{System:}\\
		\texttt{You are the Curator module of MineEvolve. Validate candidate skill/remedy entries}\\
		\texttt{and select the most relevant entries for the current planning context. Reject entries}\\
		\texttt{that are incomplete, not matchable to state fields, non-executable, overly generic,}\\
		\texttt{or conflicting with high-confidence knowledge. Output JSON only.}\\[0.5ex]
		\texttt{User:}\\
		\texttt{Current task goal: craft iron pickaxe}\\
		\texttt{Current state: inventory=\{"iron\_ingot":3,"stick":2\}, coords=[100,64,200]}\\
		\texttt{Recent failure types: []}\\
		\texttt{Prompt budget: 512 tokens}\\
		\texttt{Candidate knowledge entries: [ ... ]}\\
		\texttt{Existing knowledge store summary: [ ... ]}\\[0.5ex]
		\texttt{Output schema:}\\
		\texttt{\{}\\
		\quad\texttt{"accepted": ["skill\_id\_1", "remedy\_id\_1"],}\\
		\quad\texttt{"rejected": [}\\
		\quad\quad\texttt{\{"id":"...", "reason":"schema|match|exec|specificity|conflict"\}}\\
		\quad\texttt{],}\\
		\quad\texttt{"retrieved\_skills": ["skill\_id\_1"],}\\
		\quad\texttt{"retrieved\_remedies": ["remedy\_id\_1"],}\\
		\quad\texttt{"budget\_used": 0}\\
		\texttt{\}}
	}%
}

\newpage
\paragraph{Adaptor: knowledge-guided local plan repair.}
Adaptor is invoked when repeated failures or stagnation occur. It preserves the completed or still-valid plan prefix and regenerates only the unfinished suffix under retrieved skills and active remedies.

\noindent\framebox[\linewidth][l]{%
	\parbox{0.97\linewidth}{\small
		\textbf{Adaptor local repair prompt.}\\[0.5ex]
		\texttt{System:}\\
		\texttt{You are the Adaptor module of MineEvolve. Repair only the unfinished suffix of}\\
		\texttt{the current plan. Preserve the completed or still-valid prefix. Use retrieved skills}\\
		\texttt{as positive guidance and active remedies as repair constraints. Output JSON only.}\\[0.5ex]
		\texttt{User:}\\
		\texttt{Task goal: collect oak log}\\
		\texttt{Current state: inventory=\{\}, coords=[118,64,208]}\\
		\texttt{Current plan:}\\
		\texttt{\ \ - sg\_001: move to oak tree [failed]}\\
		\texttt{\ \ - sg\_002: mine oak log [unfinished]}\\
		\texttt{Frozen prefix: []}\\
		\texttt{Recent typed feedback:}\\
		\texttt{\ \ - e\_121: y=0, f="NAV\_STUCK", p=0.05, stagnant=true, delta\_inv=\{\}}\\
		\texttt{Retrieved skills: []}\\
		\texttt{Active remedies:}\\
		\texttt{\ \ - remedy\_nav\_obstacle: if navigation is stagnant with no inventory gain,}\\
		\texttt{\ \ \ \ clear blocking blocks or choose an alternative route before retrying.}\\[0.5ex]
		\texttt{Output schema:}\\
		\texttt{\{}\\
		\quad\texttt{"plan\_id": "p\_repair\_xxxx",}\\
		\quad\texttt{"kept\_prefix": [],}\\
		\quad\texttt{"repaired\_suffix": [}\\
		\quad\quad\texttt{\{}\\
		\quad\quad\quad\texttt{"subgoal\_id": "sg\_001r",}\\
		\quad\quad\quad\texttt{"condition": "clear blocking dirt",}\\
		\quad\quad\quad\texttt{"task\_kind": "mine",}\\
		\quad\quad\quad\texttt{"executor\_hint": "stevei",}\\
		\quad\quad\quad\texttt{"mode": "stay",}\\
		\quad\quad\quad\texttt{"timeout\_s": 45,}\\
		\quad\quad\quad\texttt{"checks": [\{"type":"path\_clear"\}],}\\
		\quad\quad\quad\texttt{"repair\_source": "remedy\_nav\_obstacle"}\\
		\quad\quad\texttt{\}}\\
		\quad\texttt{],}\\
		\quad\texttt{"active\_remedies\_used": ["remedy\_nav\_obstacle"]}\\
		\texttt{\}}
	}%
}

\end{document}